%% file: fair.tex
\documentclass[twocolumn]{fairmeta}

\usepackage{microtype}
\usepackage{graphicx}
\usepackage{booktabs} 

\usepackage{hyperref}

\usepackage{amsmath}
\usepackage{amssymb}
\usepackage{mathtools}
\usepackage{amsthm}
\usepackage{bm}

\usepackage{subcaption}
\usepackage{caption}

\usepackage{colortbl}
\usepackage{xcolor}
\usepackage{times}

\newcommand{\method}{Llip}	

\definecolor{llip}{HTML}{d9fae6}

\input{math_commands.tex}

\title{Modeling Caption Diversity in Contrastive Vision-Language Pretraining}

\author[1,2]{Samuel Lavoie}
\author[1,3,*]{Polina Kirichenko}
\author[1,*]{Mark Ibrahim}
\author[1]{Mahmoud Assran} 
\author[3]{\\Andrew Gordon Wilson}
\author[2,4]{Aaron Courville}
\author[1]{Nicolas Ballas}

\affiliation[1]{FAIR at Meta}
\affiliation[2]{Mila, Université de Montréal}
\affiliation[3]{New York University}
\affiliation[4]{Cifar fellow}

\contribution[*]{Equal contribution}

\abstract{
There are a thousand ways to caption an image.
Contrastive Language Pretraining (CLIP) on the other hand, works by mapping an image and its caption to a single vector---limiting how well CLIP-like models can represent the diverse ways to describe an image. 
In this work, we introduce Llip, Latent Language Image Pretraining, which models the diversity of captions that could match an image.
Llip's vision encoder outputs a set of visual features that are mixed into a final representation by conditioning on information derived from the text.
We show that Llip outperforms non-contextualized baselines like CLIP and SigLIP on a variety of tasks even with large-scale encoders. Llip improves zero-shot classification by an average of $2.9\%$ zero-shot classification benchmarks with a ViT-G/14 encoder. Specifically, Llip attains a zero-shot top-1 accuracy of $83.5\%$ on ImageNet outperforming a similarly sized CLIP by $1.4\%$. We also demonstrate improvement on zero-shot retrieval on MS-COCO by $6.0\%$.
We provide a comprehensive analysis of the components introduced by the method and demonstrate that Llip leads to
richer visual representations.
}

\date{\today}
\correspondence{Samuel Lavoie at \email{samuel.lavoie.m@gmail.com}}
\metadata[Code]{\href{https://github.com/facebookresearch/Llip}{https://github.com/facebookresearch/Llip}}

\begin{document}

\maketitle

\begin{figure*}[ht]
     \begin{subfigure}[b]{0.56\textwidth}
         \centering
         \includegraphics[width=\textwidth]{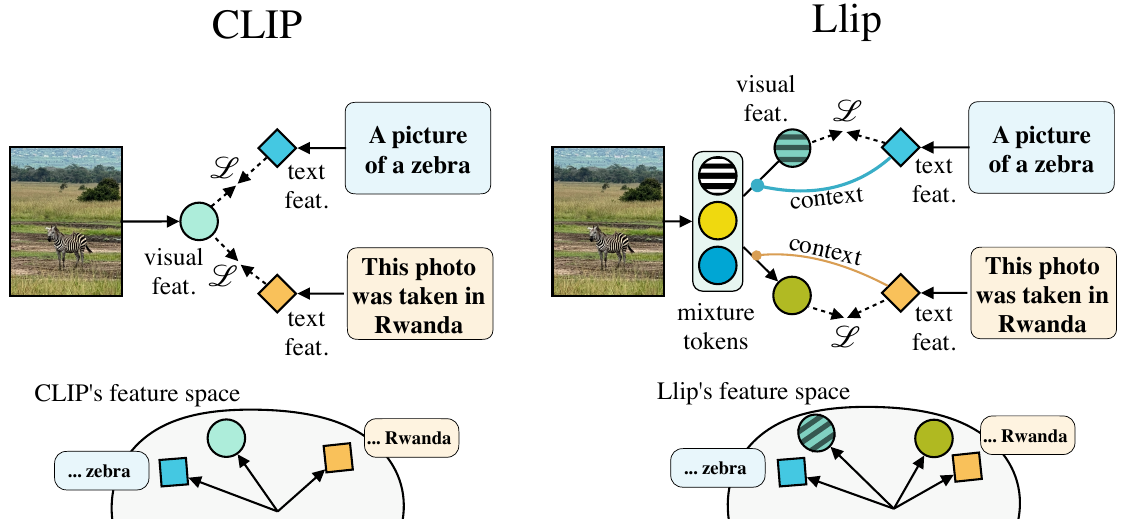}
         \caption{CLIP and Llip representations}
         \label{fig:motivation}
     \end{subfigure}
     \hfill
     \begin{subfigure}{0.4\textwidth}
     \includegraphics[width=\textwidth]{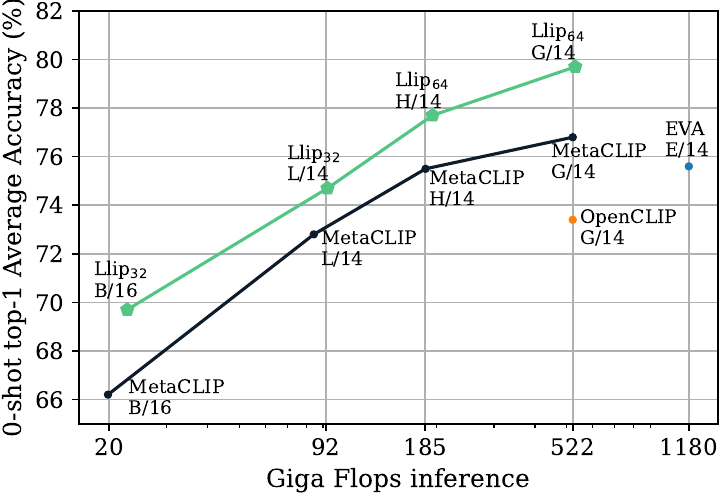}
     \caption{Averaged 0-shot classification accuracy}
     \label{fig:scale}
     \end{subfigure}
     \caption{\textbf{We propose Llip, Latent Language Image Pretraining, to model the diversity of matching captions for a given image.
     } \mbox{\textbf{(a)}~Conceptual} visualization of CLIP (left) and Llip (right) architectures.
     CLIP independently encodes visual features (shown in circles) and text features (shown in squares) which are pulled closer together by maximizing the cosine similarity objective $\mathcal{L}$. The single image feature vector of CLIP has to compromise between all matching text features (illustrated in the feature manifold at the bottom of the Figure).
     Llip outputs a set of \textit{visual mixture tokens} which are combined into a final visual feature vector conditioned on \textit{the context} derived from the caption. Llip's visual representations can more accurately represent each caption.
     \mbox{\textbf{(b)}~Zero-shot top-1 transfer accuracy} averaged over 22 established classification benchmarks (see section~\ref{sec-zshclass}) against Giga FLOPs for inference (estimated on the ImageNet zero-shot classification task)
     for encoders of various sizes. Llip  outperforms the Visual Language Pretraining baselines. Llip was trained on the same data as MetaCLIP~\citep{metaclip}. 
     }
     \label{fig:head}
 \end{figure*}

\input{introduction}

\input{method}

\input{experiments}
\input{conclusion}

\section*{Acknowledgements}
The authors want to thank Diane Bouchacourt, Florian Bordes, Hu Xu, Pascal Vincent, Pietro Astolfi, Oscar Mañas and Micah Goldblum for insightful discussions. Aaron Courville acknowledges the Canadian Research Chair and the Cifar Canadian AI Chair for their support. Polina Kirichenko and Andrew Gordon Wilson acknowledge support from NSF HDR-2118310, CDS\&E-MSS 2134216,
CAREER IIS-2145492, I-DISRE 19347.

\bibliography{ref}
\bibliographystyle{icml2024}



\clearpage
\newpage

\onecolumn
\input{appendix}

\end{document}

%% file: math_commands.tex
\usepackage{amsmath,amsfonts,bm}

\def\eqref#1{equation~\ref{#1}}

\def\1{\bm{1}}

\def\vh{{\bm{h}}}

\DeclareMathAlphabet{\mathsfit}{\encodingdefault}{\sfdefault}{m}{sl}
\SetMathAlphabet{\mathsfit}{bold}{\encodingdefault}{\sfdefault}{bx}{n}



%% file: introduction.tex
\section{Introduction}
\label{introduction}
Contrastive Language-Image Pre-training (CLIP; ~\citet{clip}) combined with a large-scale weakly supervised dataset has become the standard Visual Language Pre-training (VLP) approach to learn visual representation~\citep{li2021align,flip,sun2023evaclip,siglip,metaclip}. Due to its generality, CLIP representations are now used for many downstream tasks such as zero-shot classification~\citep{clip}, image generation~\citep{ramesh2021zero} and visual question answering~\mbox{\citep{ li2023blip,moon2023anymal}.}

At its core, CLIP aims to learn an image representation that is invariant to the caption diversity (see Figure~\ref{fig:motivation}).
CLIP uses a visual encoder and a text encoder to independently map visual and text inputs into a common representation space.  The joint encoders are trained with a contrastive objective that maximizes the similarity of representations extracted from the same image-text pair while pushing away the representations from other examples~\citep{clip}.
This training criterion encourages the representation of an image to exactly match the representation of its corresponding text description. 
Further, if different text descriptions are associated with an image, CLIP contrastive objective will push both text representations toward the same visual representation.

Yet, there is an information imbalance between the visual and text modality as visual content is often more rich than its text description~\citep{foucault1990mots}. 
Multiple diverse text captions can be equally valid descriptions of a given image, each one focusing on a different visual aspect.
For example, depending on context, someone could describe the animal from the image shown in Figure~\ref{fig:motivation} while another person could instead highlight the location where the picture was taken. Both are valid descriptions of the image and, arguably, different descriptions may capture different visual properties of the image. A training objective of a vision-language model should therefore aim at capturing the diversity of possible text descriptions to model the richness of the visual input.


In this work, we propose to explicitly model the fact that many different captions, and therefore representations, are plausible for a given image. 
To enable the prediction of different representations from a fixed image, we implement the image to text representation function as a one-to-many mapping. Conceptually, we augment our visual encoder with a latent variable that captures contextual information. Given this extra conditioning, our visual encoder can output different representations for different contexts. In our approach, the contextual latent is inferred directly from the target caption, which is then used to modulate the visual representation.

Specifically, our visual encoder is implemented by a visual transformer
that outputs $K$ learnable mixture tokens in addition to the visual tokens. The goal of the mixture tokens is to capture the different visual aspects of an input. We then make use of a cross-attention mechanism that infers the mixture token weights as a function of the text caption. The weighted mixture defines our contextual representation that is contrasted with text representations. We show that this simple modification of CLIP leads to significant improvement of the visual representation quality as illustrated in Figure~\ref{fig:scale} as well as a more rich visual representation (see Figure ~\ref{fig:evals}). 
We refer to our approach as \mbox{Latent Language Image Pre-training (\method{}).}

To demonstrate the value of our approach, we pretrain a family of vision transformer (ViT) encoders~\citep{dosovitskiy2020image} on the recent MetaCLIP~\citep{metaclip} dataset and compare our approach on various zero-shot classification and text retrieval tasks. Through an empirical evaluation and control experiments we found that:
\begin{itemize}
\item On zero-shot transfer classification, \method{} consistently outperforms CLIP pretraining for architecture of similar size on a large set of benchmarks. In particular, a VIT-G/14 encoder trained with \method{} achieves a \mbox{top-1} accuracy of 83.5\% on the ImageNet 0-shot task outperforming a VIT-G/14 trained with CLIP by 1.4\%.
\item On zero-shot image-text and text-image retrieval, \method{} consistently outperforms CLIP pretraining on COCO by 6.0\% image-to-text retrieval. 
\end{itemize}

%% file: method.tex
\begin{figure*}
     \centering
        \hfill
     \begin{subfigure}[t]{0.4\textwidth}
         \centering
         \includegraphics[width=\textwidth]{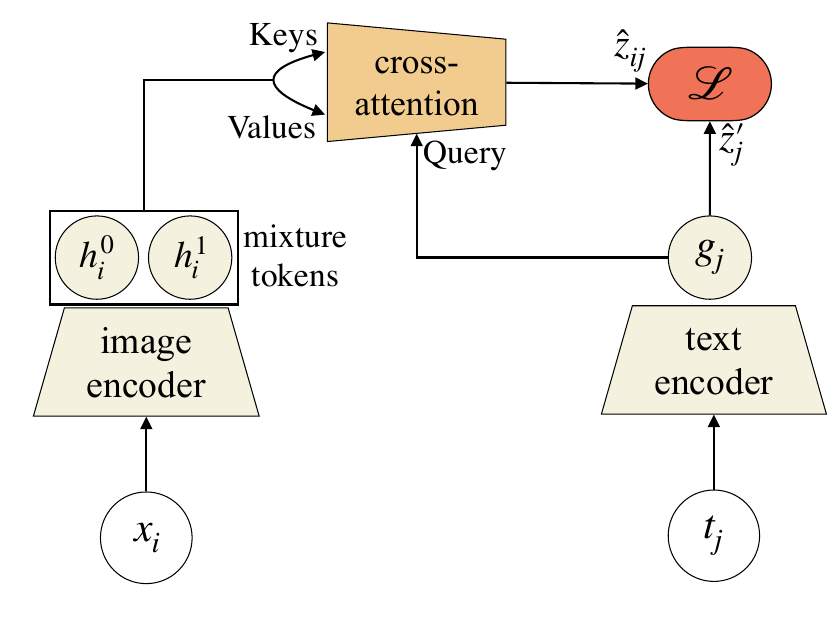}
         \caption{Llip encodes an image contextualized on the text features to compute the objective.}
         \label{fig:contextualized_contrastive}
     \end{subfigure}
     \hfill
     \begin{subfigure}[t]{0.5\textwidth}
         \centering
         \includegraphics[width=\textwidth]{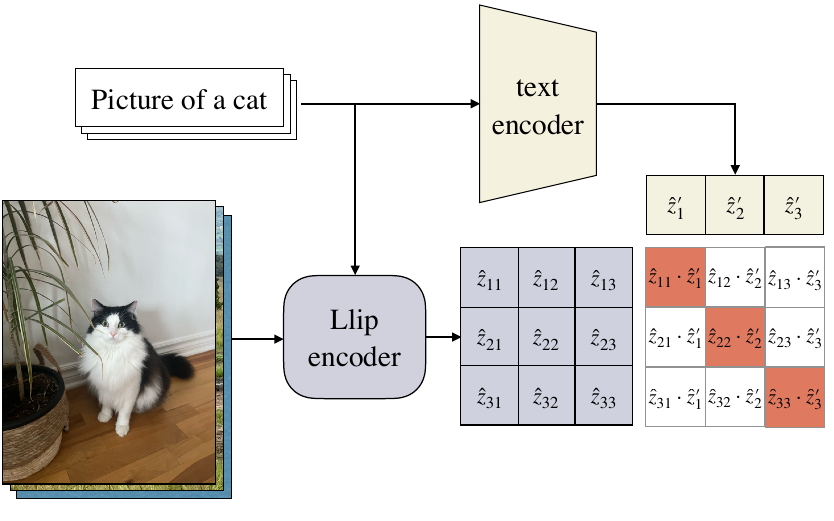}
         \caption{Training~\method{} requires encoding an image with the target text caption.}
         \label{fig:cmp_training}
     \end{subfigure}
        \caption{\textbf{Summary of the method~\method{}.} (a) Schema of \method{}'s computation of the loss. An image encoder outputs $K$ \emph{mixture tokens} ($K=2$ in the schema). The mixture tokens are given to a cross-attention module as keys and values along with the text encoding that is given as the query. The visual representation to be contrasted with the text target is conditioned on the text itself, allowing the model to produce a different visual representation depending on the caption. (b) ~\method{} uses a contrastive objective and requires encoding the visual representation with the text targets to compute the loss.}
        \label{fig:cmp_summary}
\end{figure*}

\section{Related work}


\textbf{Invariant representation.}  Invariance-based representation learning such as contrastive approaches aims at learning encoders that map two related inputs to the same point in representation space.  This paradigm is commonly used in self-supervised learning (SSL)  using a joint-embedding architecture~\citep{bromley1993signature} where the two related inputs are two transformations of the same image~\citep{demystifying2020,pirl2020,simclr}. In this case, the goal is to learn an invariant representation to a set of predefined image transformations that preserve the semantic content of the images~\citep{simclr,assran2022masked,demystifying2020,pirl2020,simclr,oquab2023dinov2}. 
While SSL methods can choose which invariance to promote through the choice of the transformations, it is not the case in vision-language pretraining as the two inputs of the encoders are from different modalities, i.e.\ an image and its text description. We hypothesize that enforcing invariance between image and text is not a desirable training objective as many text descriptions, capturing different visual aspects, could correspond to a given image.

\textbf{Predictive representation.}
Another line of works in SSL learns  representation without relying on invariant loss with the use of a joint-embedding predictive architectures (JEPA)~\citep{lecun2022path, baevski2022data2vec, assran2023self,bardes2024revisiting}.
Given a pair of related inputs $x$ and $t$, JEPA approaches learn by predicting the representation of $t$ from $x$ conditioned on a context variable that indicates the transformation between $x$ and $t$. In practice,  this idea has been explored in mask-modeling formulation where the conditioning indicates the position of $t$~\citep{baevski2022data2vec, assran2023self}.
Our approach~\method{} uses a similar learning principle in the context of vision-language pretraining. Our goal is to predict a text representation from the image input (see Figure~\ref{fig:contextualized_contrastive}).
One key difference with previous works is that we don't have a direct access to the conditioning variable which specifies the relative transformation from an image to its caption, \method{} has to infer it using the text description.

\textbf{Vision-Language Pretraining.}
A wide variety of prior works explored vision-language pretraining. ~\citet{jia2021scaling,openclip,li2023clipav2,sun2023evaclip,siglip,fini2023improved,mu2021slip} propose alternative contrastive-based Vision-Language Pretraining methods.
Some VLP methods incorporate frozen feature extractors for image or text encoders~\citep{zhai2022lit,li2023blip2,moayeri2023texttoconcept}.
Other approaches use instruction tuning \citep{liu2023improved}, context \citep{Zhou_2022_CVPR}, and grounding objectives \citep{Zhang_2021_CVPR,li2022grounded, fiber} that require additional training data for supervision. 
\citet{gao2022pyramidclip, desai2024hyperbolic} tackle the lack of a one-to-one-correspondence between web-crawled images and captions by incorporating a hierarchical loss.
All these prior works 
encourage invariance between image and text. 
Beyond contrastive pretraining, \citet{wang2022simvlm,wang2022ofa,yu2022coca,blip,blip2,fiber} incorporate a decoder with a captioning loss into vision-language models in addition to the contrastive objective.
\citet{chen2020uniter, li2021align, li2020unimo, blip} among others use an early or hybrid fusion of visual and text features using vision-grounded text encoder, i.e.\ cross-attention layers in the text encoder that attend to the output image patch tokens, which improves performance on downstream tasks but comes at a significantly increased computation cost.
In our work we instead only apply a cross-attention operation to the output of vision and text encoders, and use it to mix the final visual representation vector from the mixing tokens and context inferred from the caption.
%
In general, our approach is different from previous works in that it learns to model the diverse captions for an image solely with a contrastive objective. 



\section{Latent Language Image Pre-training}
\label{method:CMP}

This section describes our proposed method: Latent Language Image Pretraining.
\method{} learns to output a visual representation that is conditioned on a text caption. Thus, an image have a different representation depending on the caption considered during the inference.
Our approach relies on two architectural components (see Figure~\ref{fig:cmp_summary}): a visual encoder that outputs $K$ visual mixtures components, and a cross-attention module that selects how to weight the different mixture components based on the text representation.


\textbf{Visual mixture tokens.}
The image encoder is parameterized as a Vision Transformer (ViT)~\citep{dosovitskiy2020image} which processes $K$ learnable tokens along with each patch of the image~\citep{darcet2023vision}. Those learnable tokens are referred as the \textit{visual mixture tokens}.
The parameterization of our text encoder  follows
the CLIP's text encoder~\citep{clip} and outputs a single vector representation.

\textbf{Contextualization.}
\method{} conditions the visual representation using the text representation through a multi-head cross-attention mechanism.

Let $(x_i, t_i)$ be an image and a text caption from a dataset. We assume that $x_i$ and $t_j$ are a positive pair if $i=j$. Otherwise, they are a negative pair.
An image encoder $x_i \mapsto \vh_{i}$ maps an image to $K$ visual mixture tokens $\vh_i$ with $h_i^k$ for $k\in[K]$ being
the $k^{th}$ mixture tokens.
A text encoder $t_j \mapsto g_j$ maps a caption to a text feature vector. 

We denote the index of each head of a multi-head cross-attention module as $m\in[M]$. The cross-attention queries are a~projection of the text representation $g_j$: $\mathcal Q^m_j := g_j\cdot W_{\mathcal Q}^m$.
The cross-attention keys and values are the projections of the visual mixture tokens: $\mathcal K^{mk}_i := h_i^k\cdot W_{\mathcal K}^{mk}$ and $\mathcal V^{mk}_i := h_i^k \cdot W_{\mathcal V}^{mk}$. The keys, queries and values of the attention are all vectors in $\mathbb{R}^{D/M}$ as defined in~\citet{vaswani2023attention}. The mixing weights for head $m$ are defined as:
\begin{equation}
    \Phi^{m}_{ij} := \sigma_\tau ((\mathcal Q^m_j \cdot \mathcal K^{mk}_i)_{k=1}^K),
\end{equation}
with $\sigma_\tau$ being a softmax with temperature $\tau$ computed over the $K$ mixture tokens: $\sigma_\tau(z) := \dfrac{e^{z_{k}/\tau}}{\sum_{i=1}^K e^{z_i/\tau}}~\forall k\in[K]$.
From the mixing weights and $\mathcal V$, we compute the contextualized visual representation:
\begin{equation}\label{eqn:mixing}
    z_{ij} := \text{Concat}\left(\left(\sum_{k=1}^K {\Phi^{mk}_{ij} \cdot \mathcal V^{mk}_{i}}\right)_{m=1}^M\right) \cdot W_\mathcal{O},
\end{equation}
where $W_\mathcal{O}$ is a learnable projection matrix in $\mathbb{R}^{D \times D}$. 

Similarly we project the text representation  ${z_j':=g_j^t\cdot W_T}$ where $W_T$ is learnable projection matrix of the text features. Both representation are normalized as previously done in CLIP when computing the objective function: $\hat{z}_{ij} = \dfrac{z_{ij}}{||z_{ij}||_2}$ and $\hat{z}'_j=\dfrac{z'_j}{||z'_j||_2}$.

\textbf{Pretraining.}
For pretraining, we consider the SigLIP \citep{siglip} objective due to its memory efficiency.
We modify SigLIP's objective using our contextualized visual representation and propose the following loss:
\begin{equation}
    \begin{split}
        \mathcal{L}_\text{\method{}} := \dfrac{1}{N}\sum_{i=1}^N &\log\dfrac{1}{1+e^{(-a\hat{z}_{ii}\cdot \hat{z}_{i}' + b)}} +
            \\ \dfrac{1}{N}\sum_{i=1}^N\sum_{j=1;i\neq j}^N &\log\dfrac{1}{1+e^{(a\hat{z}_{ij}\cdot \hat{z}_{j}' - b)}},
\end{split}
\label{eqn:illip}
\end{equation}
where $a$ and $b$ are learnable parameters, $N$ is the size of the mini-batch, $\hat z_j'$ is the text representation obtained from caption $j$ and $\hat z_{ij}$ is the visual representation obtained from mixing the visual mixture tokens of image $i$ with the text features of caption $j$.

\textbf{Avoiding a shortcut solution.}
Contextualizing the visual features with the target caption can introduce a shortcut solution: the network ignores $x_i$ and solely relies on $t_i$ to minimize its objective. The negative samples of the contrastive objective in~\eqref{eqn:illip} prevent that shortcut solution. While, the caption $t_i$ is a positive caption for $x_i$, the same caption is also a negative caption for a different sample $x_j$. Therefore, relying only on $t_i$ is not a valid solution because the objective also minimizes the similarity for pairs of negative samples, i.e. it pushes away $\hat z_{ji}$ from $\hat{z}_j$. 


\textbf{Inference.}
The final visual representation depends on a caption.
Consequently each image has to be encoded with all target captions as illustrated in Figure~\ref{fig:cmp_training}, both for pre-training and zero-shot evaluation.
Fortunately, the fusion of the image and text is lightweight as it occurs in the output layer. The additional compute and memory cost is constant for a fixed number of mixture tokens $K$ as we scale up the size of the encoder (See Figure~\ref{fig:scale_inference_time}).

Inference for zero-shot classification in Llip is analogous to CLIP’s implementation. For a given image $x_i$, we have $C$ possible caption labels $t_j, j \in [C]$. 
We encode each image $x_i$ with each caption label $t_j$ obtaining contextualized visual features $z_{ij}$.
Then we compute the cosine similarity between the normalized visual features $\hat z_{ij}$ and text features $\hat z_j'$, and define the predicted label as the one with the highest cosine similarity between the contextualized image features and the text features.

%% file: experiments.tex
\section{Experimental Setup}
\label{experiments}

Our empirical analysis over the next sections has three main objectives. First, we aim to demonstrate the contribution of each modification added by~\method{} via controlled experiments.
Second, we illustrate the value of~\method{} in comparison to other contrastive VLP methods on a set of standard zero-shot benchmarks commonly used in the literature. 
Finally, we provide an comprehensive analysis of~\method{} representations and hyper-parameters. Before discussing our results, we describe our experimental  setup.

We perform our experiments on 5 models: \mbox{ViT-B/32}, \mbox{ViT-B/16}, \mbox{ViT-L/14}, \mbox{ViT-H/14} and \mbox{ViT-G/14}. ViT-B/32 stands for a base Vision Transformer with image patch of size 32 and ViT-L/14 is a large Vision Transformer with patch of size 14 (see~\citet{dosovitskiy2020image} for implementation details).
To capture the visual variability in images, our method appends $K$ additional  learnable tokens to the input sequence of transformers, similarly to~\citet{darcet2023vision}. We refer to those extra tokens as mixture tokens and we denote the model with $K$ mixture tokens by $\text{~\method{}}_K$.
For all of our experiments, we crop and resize images to $224\times 224$.

We pre-train our models with the AdamW optimizer~\citep{kingma2017adam, loshchilov2017decoupled} with $\beta_2 = 0.95$ as done by~\citet{siglip} to stabilize the pre-training. We use a learnable scale parameter $a$ along with a learnable bias $b$ for our objective following the initialization of~\citet{siglip}.
Otherwise, all other training decisions closely follow the ones used by~\citet{clip,metaclip}.
For all of the~\method{} experiments, we fix $M=8$ the number of heads in the cross-attention.
Unless mentioned otherwise, the cross-attention's temperature $\tau=5$.

Our models were trained on the Common Crawl data curated using the methodology presented in~\citet{metaclip}. We use a dataset of 2.5B image-text pairs collected using the same parameters that was used in~\citet{metaclip}. 
As done in~\citet{clip,metaclip} we pre-train our model for a total amount of 12.8B pairs of image-text seen with a batch size of 32,768.

To increase the training efficiency, we leverage compilation and mixed-precision in PyTorch~\citep{pytorch}. We use gradient checkpointing for computing the activations of the visual representations to reduce the memory during pre-training. The ViT-B and ViT-L models were trained on 128 V100 and A100 respectively. The larger models were trained on 256 A100 80GB GPUs. 


\section{From SigLIP to~\method{}}

\label{sec:buildup}
\begin{figure}
    \centering
    \includegraphics[width=0.45\textwidth]{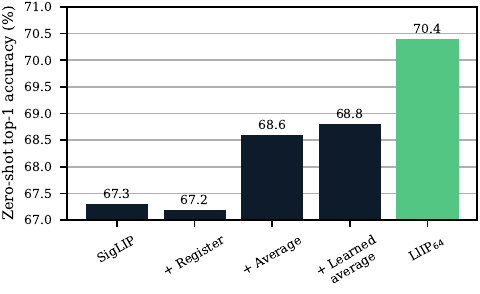}
    \caption{\textbf{Decomposing the effects of Llip's ingredients.} Ablation of the added components of \method{} compared to SigLIP and their effect on zero-shot ImageNet transfer accuracy. Every models are trained with a ViT-B/32. From left to right, we evaluate:
    1) Re-implemented SigLIP baseline, 2) adding additional $63$ mixture tokens (+Registers \citep{darcet2023vision}) which are not used in the final representation,
    3) using uniform mixing of the learnable tokens (+Average), 4) non-uniform mixing of the tokens (+Learned average), 5) context-conditional mixing of the tokens (\method{}$_{64}$). Conditioning the mixing weights of the tokens on the text feature achieves the best performance.}
    \label{fig:building_lalip}
\end{figure}

To assess the impact of the contextualization of \method{}, we explore how the performance evolves when gradually modifying an existing SigLIP baseline toward \method{}. 
Our starting baseline SigLIP pre-training with a ViT-B/32 and  the MetaCLIP dataset. We introduce three intermediate baselines -- each corresponding to an intervention on the previous baseline -- that gradually interpolate between SigLIP and \method{} in the way the visual representation is computed. We present their respective performances on ImageNet zero-shot top-1 accuracy in Figure~\ref{fig:building_lalip}.

\textbf{SigLIP.~}
We reproduce SigLIP pre-training with our setup. The zero-shot accuracy on ImageNet is similar to the accuracy of 67.6 reported by MetaCLIP~\citep{metaclip}.

\textbf{+ Register.~}
We increase the amount of learned tokens from $1$ to $64$ in SigLIP, but only use the first learned token to compute SigLIP objective as done in~\citet{darcet2023vision} (they refer to additional tokens as registers). This procedure does not improve the ImageNet top-1 accuracy.

\textbf{+ Average.} Next, we explore the effect of tokens mixing.
 We compute equal-weighted average of all of the $64$ learned tokens and use the resulting vector to compute the objective.
 We find that averaging the learned tokens leads to a significant improvement over the baseline. Adding extra learned tokens and uniform mixing is an effective method to improve VLP.

\textbf{+ Learned Average.}
We introduce non-uniform mixing to aggregate the mixture tokens.
We apply a cross-attention operation as described in~\eqref{eqn:mixing} except the query is a learned vector shared across all samples instead of the text caption.
We don't find a significant difference between uniform and non-uniform mixing of the learned tokens.

\textbf{\method{}.} Finally, we contrast the aforementioned baselines with \method{} where the mixing weights now depend on the text features, i.e.\ the query token for the cross attention is a function of the text representation. \method{} shows significant improvement over the average baseline in zero-shot Top-1 ImageNet accuracy.

We find that strong performance of~\method{} comes from mixing visual features conditioned on the text features. 

\begin{table*}[ht]
\centering
\scriptsize
\setlength{\tabcolsep}{2.5pt}
\caption{\textbf{Zero-shot classification benchmarks when pretraining on the MetaCLIP dataset} on ViT-B/32, ViT-B/16, ViT-L/14, ViT-H/14 and ViT-G/14. We compare Llip to CLIP and SigLIP for several backbones with different scales. We pre-train all the models with the MetaCLIP dataset and use the same pre-training recipe.
Llip  outperforms MetaCLIP across most benchmarks.
$^*$: Denotes that we reproduced the baseline with our setup. MetaCLIP numbers are reported from: $^1$:~\citep{metaclip}.}
\label{tab:zs-all-benchmarks}
\begin{tabular}{l|c|cccccccccccccc|ccccc|ccc}
&& \multicolumn{14}{c}{Standard vision} & \multicolumn{5}{c}{Fine-grained} & \multicolumn{3}{c}{Other} \\
 & \rotatebox{90}{\textbf{Average}} & \rotatebox{90}{ImageNet} & \rotatebox{90}{Food-101}& \rotatebox{90}{CIFAR10}& \rotatebox{90}{CIFAR100}& \rotatebox{90}{CUB}& \rotatebox{90}{SUN397}& \rotatebox{90}{Cars}& \rotatebox{90}{Aircraft}& \rotatebox{90}{Pets}& \rotatebox{90}{Caltech-101}& \rotatebox{90}{Flowers}&\rotatebox{90}{MNIST}&\rotatebox{90}{STL-10}& \rotatebox{90}{GTSRB}&
 \rotatebox{90}{DTD}& \rotatebox{90}{EuroSAT}& \rotatebox{90}{RESISC45}&  \rotatebox{90}{PCAM}& \rotatebox{90}{Country211}& \rotatebox{90}{KITTI}&  \rotatebox{90}{UCF101}& \rotatebox{90}{MIT-States}\\
\hline
\textit{ViT-B/32} &&&&&&&&&&&&&&&&&&&&&&& \\
MetaCLIP$^1$ & 62.8 & 67.6 & 82.7 & 95.2 & 77.7 & 67.8 & 66.8 & 77.4 & 27.0 & 90.9 & 92.8 & 69.9 & 42.7 & 96.3 & 39.2 & \textbf{58.9} & 51.1 & 66.3 & 50.0 & 17.7 & 29.3 & 67.5 & 47.6 \\
SigLIP$^*$ & 63.5 & 67.3 & 81.8 & 94.8 & 77.1 & 68.9 & 66.5 & 78.7 & 29.0 & 88.9 & \textbf{93.0} & 70.3 & 41.9 & 96.8 & 52.3 & 58.8 & 47.4 & 64.7 & 54.8 & 17.0 & 30.9 & 69.5 & 46.9  \\
\rowcolor{llip}
$\text{\method{}}_{64}$ & \textbf{67.5} & \textbf{70.4} & \textbf{84.1} & \textbf{95.5} & \textbf{80.8} & \textbf{71.5} & \textbf{68.6} & \textbf{82.2} & \textbf{34.9} & \textbf{92.3} & 92.9 & \textbf{74.8} & \textbf{66.3} & \textbf{97.5} & \textbf{53.6} & 58.8 & \textbf{49.9} & \textbf{67.5} & \textbf{64.5} & \textbf{20.7} & \textbf{37.8} & \textbf{71.6} & \textbf{48.5}  \\
\hline
\textit{ViT-B/16} &&&&&&&&&&&&&&&&&&&&&&& \\
MetaCLIP$^1$ & 66.2 & 72.1 & 88.3 & 95.7 & 79.0 & 71.4 & 68.5 & 82.9 & 30.3 & 91.7 & 93.3 & 73.9 & 66.1 & 98.4 & 46.6 & 62.1 & 51.1 & 71.1 & 50.5 & 22.7 & 16.6 & 73.0 & 50.4 \\
SigLIP$^*$ & 67.1 & 72.3 & 88.5 & \textbf{96.0} & 79.0 & 74.1 & 68.5 & 83.5 & 33.8 & 92.2 & 94.2 & 72.5 & 63.3 & 98.5 & 40.8 & 60.3 & 50.1 & \textbf{68.6} & \textbf{55.5} & 22.0 & \textbf{38.2} & 74.3 & 50.4 \\
\rowcolor{llip}
$\text{\method{}}_{64}$ & \textbf{69.7} & \textbf{75.3} & \textbf{89.0} & 95.7 & \textbf{81.4} & \textbf{75.0} & \textbf{70.9} & \textbf{88.2} & \textbf{41.5} & \textbf{93.5} & \textbf{94.7} & \textbf{74.9} & \textbf{79.6} & \textbf{98.5} & \textbf{54.0} & \textbf{63.7} & \textbf{56.7} & 67.6 & 53.1 & \textbf{25.7} & 24.9 & \textbf{77.6} & \textbf{51.7} \\
\hline\textit{ViT-L/14} &&&&&&&&&&&& \\
MetaCLIP$^1$ & 72.8 & 79.2 & 93.5 & 97.6 & 84.2 & 80.1 & 73.7 & 88.7 & 44.4 & 94.7 & 95.5 & 81.8 & 64.4 & \textbf{99.3} & 56.3 & 68.3 & 58.7 & 74.6 & \textbf{66.5} & 34.0 & 29.7 & 81.7 & 55.6 \\
SigLIP$^*$ & 73.9 & 79.4 & 93.2 & 97.6 & 84.0 & \textbf{82.3} & 72.0 & 90.7 & 51.9 & 95.5 & \textbf{95.7} & \textbf{83.1} & 67.4 & 99.2 & 67.3 & 69.2 & 58.0 & 74.4 & 55.6 & 33.3 & \textbf{37.4} & 82.4 & 55.5 \\
\rowcolor{llip}
$\text{\method{}}_{32}$ & \textbf{74.7} & \textbf{80.9} & \textbf{93.6} & \textbf{98.0} & \textbf{86.8} & 81.2 & \textbf{74.4} & \textbf{91.7} & \textbf{55.1} & \textbf{96.0} & 95.2 & 81.4 & \textbf{68.0} & \textbf{99.3} & \textbf{68.8} & \textbf{69.8} & \textbf{59.8} & \textbf{77.3} & 54.7 & \textbf{36.4} & 34.8 & \textbf{84.5} & \textbf{56.1}  \\
\hline\textit{ViT-H/14} &&&&&&&&&&&&&&&&&&&&&&& \\
MetaCLIP$^1$ & 75.5 & 80.5 & 94.2 & \textbf{98.0} & 86.4 & 83.4 & 74.1 & 90.0 & 50.2 & 95.4 & 95.6 & 85.1 & 72.7 & \textbf{99.4} & 62.5 & 72.4 & \textbf{66.3} & \textbf{74.6} & 65.8 & 37.2 & \textbf{38.2} & 82.2 & 56.2 \\
\rowcolor{llip}
$\text{\method{}}_{64}$ & \textbf{77.7} & \textbf{82.7} & \textbf{95.1} & 97.9 & \textbf{87.2} & \textbf{86.2} & \textbf{75.0} & \textbf{92.4} & \textbf{61.3} & \textbf{96.0} & \textbf{95.8} & \textbf{86.4} & \textbf{86.6} & \textbf{99.4} & \textbf{70.8} & \textbf{72.8} & 62.4 & 74.2 & \textbf{68.6} & \textbf{41.3} & 33.6 & \textbf{86.2} & \textbf{57.2} \\
\hline\textit{ViT-G/14} &&&&&&&&&&&&&&&&&&&&&&& \\
MetaCLIP$^1$ & 76.8 & 82.1 & 94.9 & \textbf{98.5} & 88.6 & 84.0 & 74.7 & 90.9 & 52.7 & 96.1 & 95.7 & \textbf{89.5} & 78.1 & \textbf{99.5} & 61.6 & 72.6 & \textbf{73.7} & 75.5 & 65.6 & 41.5 & 31.0 & 85.6 & 56.6 \\
\rowcolor{llip}
$\text{\method{}}_{64}$ & \textbf{79.7} & \textbf{83.5} & \textbf{95.6} & \textbf{98.5} & \textbf{89.5} & \textbf{86.8} & \textbf{76.5} & \textbf{93.6} & \textbf{67.4} & \textbf{96.7} & \textbf{95.8} & \textbf{89.5} & \textbf{89.9} & \textbf{99.5} & \textbf{72.5} & \textbf{75.7} & 70.7 & \textbf{77.7} & \textbf{71.9} & \textbf{45.6} & \textbf{31.1} & \textbf{88.0} & \textbf{57.9} \\
\hline
\end{tabular}
\end{table*}

\begin{table*}[ht]
\centering
\caption{\textbf{Zero-shot retrieval on Flickr30k~\citep{flickr} and MSCOCO~\citep{mscoco}.} Comparison of zero-shot retrieval performances of \method{} with the SigLIP and MetaCLIP baselines. All methods are pre-trained with the same dataset and use the same pre-training recipe. We compare both Image to Text and Text to Image retrievals. \method{} demonstrate consistent gain for both MSCOCO and Flicker30k. $^*$: Reproduced number with our setup. MetaCLIP results are  reported from: 
$^1$:~\citep{metaclip}.
}
\label{tab:zs-retrieval}
\scriptsize
\begin{tabular}{lcccccc|cccccc}
& \multicolumn{6}{c}{Image$\to$Text} & \multicolumn{6}{c}{Text$\to$Image} \\
 & \multicolumn{3}{c}{Flickr30K} & \multicolumn{3}{c}{MSCOCO}  & \multicolumn{3}{c}{Flickr30K} & \multicolumn{3}{c}{MSCOCO} \\
 & R@1 & R@5 & R@10 & R@1 & R@5 & R@10 & R@1 & R@5 & R@10 & R@1 & R@5 & R@10 \\
\hline
\emph{ViT-B/16:} &&&&&&&&&&&&\\
MetaCLIP$^1$ & 85.9 & 97.3 & 98.9 & 59.4 & 80.6 & 87.9 & 70.5 & 90.7 & 94.6 & 41.4 & 67.2 & 77.0 \\
SigLIP$^*$ & 85.4 & 97.1 & 98.6 & 59.7 & 82.1 & 89.1 & 69.6 & 90.0 & 94.1 & 42.0 & 67.3 & 77.0 \\
\rowcolor{llip}
$\text{\method{}}_{64}$ & \textbf{90.1} & \textbf{98.5} & \textbf{99.6} & \textbf{63.4} & \textbf{84.3} & \textbf{90.3} & \textbf{75.1} & \textbf{92.8} & \textbf{96.2} & \textbf{45.6} & \textbf{70.8} & \textbf{79.7} \\
\hline
\emph{ViT-L/14:} &&&&&&&&&&&& \\
MetaCLIP$^1$ & 90.4 & 98.5 & 99.1 & 64.5 & 85.0 & 91.3 & 76.2 & 93.5 & 96.4 & 47.1 & 71.4 & 80.3 \\
SigLIP$^*$ & 91.5 & 98.1 & \textbf{99.4} & 65.4 & 85.1 & 91.1 & 76.5 & 94.3 & 96.6 & 48.1 & 72.3 & 80.6 \\
\rowcolor{llip}
$\text{\method{}}_{32}$ & \textbf{93.2} & \textbf{99.0} & \textbf{99.4} & \textbf{68.1} & \textbf{87.6} & \textbf{92.5} & \textbf{79.9} & \textbf{95.0} & \textbf{97.4} & \textbf{50.6} & \textbf{74.7} & \textbf{82.8} \\
\hline
\emph{ViT-H/14:} &&&&&&&&&&&& \\
MetaCLIP$^1$ & 91.6 & 98.6 & 99.7 & 66.2 & 86.2 & 91.9 & 78.0 & 94.6 & 96.9 & 48.8 & 73.2 & 81.4  \\
\rowcolor{llip}
$\text{\method{}}_{64}$ & \textbf{94.0} & \textbf{99.4} & \textbf{99.9} & \textbf{71.6} & \textbf{89.3} & \textbf{94.0} & \textbf{82.8} & \textbf{96.0} & \textbf{98.0} & \textbf{53.9} & \textbf{77.0} & \textbf{84.2}  \\
\hline
\emph{ViT-G/14:} &&&&&&&&&&&& \\
MetaCLIP$^1$ & 91.2 & 98.7 & 99.7 & 66.7 & 86.6 & 92.3 & 80.0 & 94.5 & 97.0 & 49.6 & 73.8 & 81.9  \\
\rowcolor{llip}
$\text{\method{}}_{64}$ & \textbf{94.8} & \textbf{99.7} & \textbf{100} & \textbf{72.7} & \textbf{90.1} & \textbf{94.4} & \textbf{82.5} & \textbf{96.0} & \textbf{97.9} & \textbf{54.2} & \textbf{77.1} & \textbf{84.5}  \\
\hline
\end{tabular}
\end{table*}

\begin{table*}[t]
\centering
\caption{\textbf{Comparison of zero-shot classification.}
We compare \method{} (\emph{ViT-G/14}) to the best reported number of EVA-CLIP (\emph{ViT-E/14}), OpenCLIP (\emph{ViT-G/14}) and MetaCLIP (\emph{ViT-G/14}) baselines on 22 classifications tasks involving object classification (e.g.\ ImageNet, CIFAR), fine-grained classification (e.g.\ Cars, Aircraft, Flowers), non-natural images (e.g.\ DTD, EuroSAT, PCAM).
\method{} obtains the best average performance across baselines and improves the best performance in $19$ out of the $22$ classification tasks.
We only consider baselines that reports  performance on the same tasks or that provide model weights.
$^1$:~\citep{sun2023evaclip};
$^2$:~\citep{cherti2023reproducible};
$^3$:~\citep{metaclip}.
}
\label{tab:zs-benchmarks}
\scriptsize
\setlength{\tabcolsep}{2.5pt}
\begin{tabular}{l|c|cccccccccccccccccccccc}
 & \rotatebox{90}{\textbf{Average}} & \rotatebox{90}{ImageNet} & \rotatebox{90}{Food-101}& \rotatebox{90}{CIFAR10}& \rotatebox{90}{CIFAR100}& \rotatebox{90}{CUB} & \rotatebox{90}{SUN397}& \rotatebox{90}{Cars}& \rotatebox{90}{Aircraft}& \rotatebox{90}{Pets}& \rotatebox{90}{Caltech-101}& \rotatebox{90}{Flowers}&\rotatebox{90}{MNIST}&\rotatebox{90}{STL-10}& \rotatebox{90}{GTSRB}&
 \rotatebox{90}{DTD}& \rotatebox{90}{EuroSAT}& \rotatebox{90}{RESISC45}&  \rotatebox{90}{PCAM}& \rotatebox{90}{Country211}& \rotatebox{90}{KITTI} & \rotatebox{90}{UCF101} & \rotatebox{90}{MIT-States}\\
\hline
\textit{ViT-E/14:} &&&&& \\
EVA-CLIP$^1$ & 75.6 & 82.0 & 94.9 & \textbf{99.3} & \textbf{93.1} & 85.8 & 75.1 & 94.6 & 54.1 & 95.8 & 90.5 & 84.5 & 74.7 & 99.0 & 67.7 & 68.2 & \textbf{75.8} & 75.6 & 63.7 & 35.7 & 12.4 & 83.1 & 56.7 \\
\hline
\textit{ViT-G/14:} &&&&&&&& \\
OpenCLIP$^2$ & 73.5 & 80.1 & 93.1 & 98.2 & 87.5 & 84.4 & 74.5 & 94.5 & 49.7 & 95.2 & 86.4 & 81.5 & 71.6 & 98.5 & 62.5 & 69.0 & 70.0 & 72.6 & 63.6 & 33.8 & 15.6 & 80.5 & 54.5 \\
MetaCLIP$^3$ & 76.8 & 82.1 & 94.9 & 98.5 & 88.6 & 84.0 & 74.7 & 90.9 & 52.7 & 96.1 & 95.7 & \textbf{89.5} & 78.1 & \textbf{99.5} & 61.6 & 72.6 & 73.7 & 75.5 & 65.6 & 41.5 & 31.0 & 85.6 & 56.6 \\
\rowcolor{llip}
$\text{\method{}}_{64}$ & \textbf{79.7} & \textbf{83.5} & \textbf{95.6} & 98.5 & 89.5 & \textbf{86.8} & \textbf{76.5} & \textbf{93.6} & \textbf{67.4} & \textbf{96.7} & \textbf{95.8} & \textbf{89.5} & \textbf{89.9} & \textbf{99.5} & \textbf{72.5} & \textbf{75.7} & 70.7 & \textbf{77.7} & \textbf{71.9} & \textbf{45.6} & \textbf{31.1} & \textbf{88.0} & \textbf{57.9} \\
\hline
\end{tabular}
\end{table*}

\section{Zero-shot Evaluations}

In this section, we evaluate the performance of \method{} on zero-shot classification and retrievals benchmarks.
We first present an apples-to-apples comparison between CLIP, SigLIP and Llip for various backbone sizes. We train all of the models with the MetaCLIP dataset and we fix the hyper-parameters to the one found in prior works~\mbox{\citep{siglip,metaclip}}. We observe that Llip consistently outperforms the baselines for every model sizes on both zero-shot classification transfer and zero-shot retrieval.

Next, we compare our approach with various baselines such as CLIP~\citep{clip}, OpenCLIP~\citep{cherti2023reproducible}, SigLIP~\citep{siglip}, MetaCLIP~\citep{metaclip}, CLIPA~\citep{li2023clipav2}, Data Filtering Network~\citep{dfn} that all implement a variant of constrastive learning and EVA-CLIP~\citep{sun2023evaclip} which combines contrastive objective with input masking.

\subsection{Llip improves zero-shot performance for a fixed pre-training setup}
\label{sec-zshclass}
In this subsection, we evaluate Llip and compare it to the CLIP and SigLIP contrastive approaches. All methods use the same training dataset.

We evaluate Llip on a wide variety of classification benchmarks. The classification benchmarks contain tasks on object classification (ImageNet \citep{recht2019imagenet},
CIFAR \citep{krizhevskyLearningMultipleLayers},
CUB \citep{CaltechUCSDBirds2002011Dataset},
Food-101 \citep{bossardFood101MiningDiscriminative2014},
STL-10 \citep{coatesAnalysisSingleLayerNetworks},
caltech-101 \citep{CaltechUCSDBirds2002011Dataset},
MNIST \citep{lecun-mnisthandwrittendigit-2010}),
~fine-grained classification (SUN397 \citep{xiaoSUNDatabaseLargescale2010}, 
\mbox{Cars \citep{krauseCollectingLargeScaleDataset}},
Aircraft \citep{majiFineGrainedVisualClassification2013}, 
Pets \citep{parkhiCatsDogs2012}, 
Flowers \citep{nilsbackAutomatedFlowerClassification2008}, 
GTRSB \citep{Stallkamp-IJCNN-2011}, 
Country211 \citep{clip}),
~non-natural images (\mbox{DTD~\citep{cimpoiDescribingTexturesWild2013}}, EuroSAT~\citep{helberEuroSATNovelDataset2019},
RESIS45~\citep{chengRemoteSensingImage2017},
PCAM~\citep{Ye2020WeaklySL})
~and video classification (KITTI~\citep{kitti}, UCF101~\citep{soomroucf101}) and attribute recognition (MIT-States~\citep{mit_states}).

In Table~\ref{tab:zs-all-benchmarks} demonstrates that Llip outperforms CLIP and SigLIP when controlling for the training data distribution.
On a ViT-B/32, Llip outperforms SigLIP by 4.7\% in average. On a ViT-G/14, Llip outperforms MetaCLIP by 2.9\% in average. 
Table~\ref{tab:zs-retrieval} also shows that Llip outperforms CLIP and SigLIP on the Flickr30k and MSCOCO  zero-shot retrieval tasks. 
Llip outperforms a CLIP based model on MSCOCO text retrieval by 4\% with a ViT-B/16 and 6\% with a ViT-G/14. Llip observes similar improvement on MSCOCO image retrieval with a gain of 4.2\% with a ViT-B/16 and 4.6\% with a ViT-G/14.

\subsection{Llip comparision with previous contrastive pre-training baselines}
We now compare Llip with previously reported numbers in the literature of contrastive visual language pre-training. While these numbers are obtained with different model architectures, training recipes and datasets, we observe that Llip is a competitive method.


\begin{figure}[ht]
    \centering
    \includegraphics[width=0.45\textwidth]{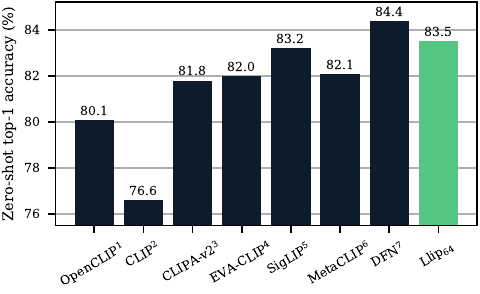}
    \caption{\textbf{ImageNet zero-shot transfer classification.}
   We compare a VIT-G/14 trained with $\text{Llip}_{64}$ with various vision-language baselines. We select the best reported number for every methods. \method{} outperforms most of the vision-language pretraining baselines on ImageNet. \method{} outperforms most of the. DFN, which is the only methods outperforming \method{}, is trained on a larger datasets of  5B curated samples and use $378$ instead of $224$ as input image resolution.
    We report the imagenet performance of the baselines from:
    $^1$:~\citep{cherti2023reproducible};
    $^2$:~\citep{clip}; 
    $^3$:~\citep{li2023clipav2}; 
    $^4$:~\citep{sun2023evaclip}; 
    $^5$:~\citep{siglip};
    $^6$:~\citep{metaclip};
    $^7$~\citep{dfn}.}
    \label{fig:sota-compare}
\end{figure}

\textbf{ImageNet.} We investigate \method{}'s zero-shot transfer performance on the ImageNet classification task \citep{russakovskyImageNetLargeScale2015}.
We report the top-1 accuracy of \method{} with a ViT-G/14 and the best reported numbers from OpenCLIP, CLIP, CLIPA-v2, SigLIP, MetaCLIP and DFN in Figure~\ref{fig:sota-compare}. 
\method{} outperforms most previous approaches. In particular, our method shows a gain +0.3\% over SigLIP while processing $4\times$ less samples during pre-training and a gain of 2.5\% over EVA-CLIP that is pre-trained with a ViT-E/14 backbone that has $2.5\times$ more parameters that the ViT-G/14. While DFN obtains a higher zero-shot top-1 accuracy than Llip,  it is trained on a larger datasets of  5B curated samples and uses $378$ instead of $224$ as input image resolution.
We conjecture that Llip may also benefit from higher quality data, but we leave such analysis to future works.

Closest in the setting of our work is MetaCLIP which trains a joint-embedding architecture using contrastive loss on the  a similar pre-training dataset. 
\method{} outperforms MetaCLIP VIT-G/14 by $+1.4\%$, highlighting the benefit of modelling the caption diversity.

\textbf{Other image classification tasks.}
To demonstrate the genericity of the learned representation with \method{}, we  measure performances across 22 standard zero-shot classification benchmarks that are usually reported in the literature in Table \ref{tab:zs-benchmarks}. 
We compare our approach with OpenCLIP, MetaCLIP and EVA-CLIP which all report results on the same set of tasks or release their model weights allowing us to evaluate and compare with these models.
Results show that \method{} obtains the best average performance across baselines. It reaches the the best performance in $19$ out of the $22$ classification tasks.



\section{Analysis of \method{}}

\begin{figure}[ht]
    \centering
    \includegraphics[width=0.45\textwidth]{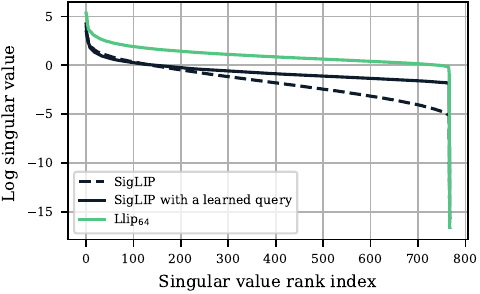}
    \caption{\textbf{\method{}'s representation is more expressive than the non-contextualized SigLIP baselines.} Singular value spectrum of the covariance matrix of the visual features of a ViT-B/32 using different pre-training objectives.
    The embedding vectors are taken at the output of the visual encoder.
     SigLIP with a learned query baseline adds $64$ mixture tokens and learns how to average them using a cross-attention with a learnable query vector. We concatenate the $64$ mixture tokens along the batch dimension for the learned query baseline and~\method{}. \method{} show slower decay in the singular value spectrum than the two baselines which indicates a larger variability of the features.}
    \label{fig:evals}
\end{figure}

\textbf{Representation expressivity.}
We evaluate the expressivity of the learned visual features by computing the singular values of the covariance matrix of the visual features as done in~\citet{jing2022understanding}.
This method was proposed to probe the dimensionality collapse in self-supervised pre-trained methods and also measures the expressiveness of learned representations~\citep{decorrelation}.

In particular, we compare SigLIP, SigLIP with learned query (see Section~\ref{sec:buildup}) and ~\method{}$_{64}$.
We collect the embedding vectors of $5000$ samples from ImageNet's validation set randomly chosen.
For SigLIP with learned query and~\method{}, we concatenate the $64$ mixture tokens along the batch dimension. Then we compute the singular value spectrum of the feature covariance matrix~\citep{jing2022understanding} that we plot in log scale in Figure~\ref{fig:evals}. \method{} show slower decay in the singular value spectrum than the two baselines which indicates a larger variability of the features. 



\begin{figure}[!ht]
    \centering
    \begin{subfigure}[t]{0.23\textwidth}
        \includegraphics[width=\textwidth]{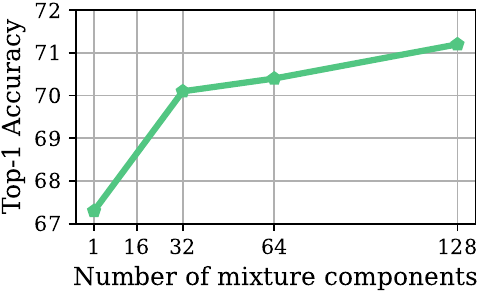}
        \caption{Number of mixture tokens.}
        \label{fig:K-lalip}
    \end{subfigure}
    \hfill
    \begin{subfigure}[t]{0.23\textwidth}
        \includegraphics[width=\textwidth]{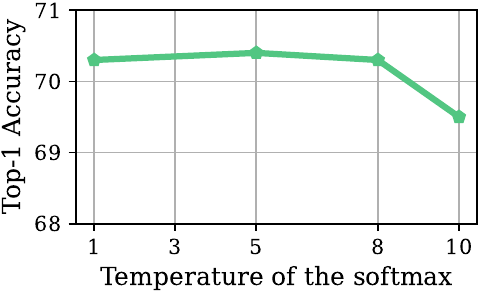}
        \caption{Attention's temperature.}
        \label{fig:t-lalip}
    \end{subfigure}
    \caption{\textbf{Analysis of~\method{}'s hyperparameters} on downstream zero-shot top-1 ImageNet accuracy for a ViT-B/32 visual encoder.
    We explore the effect of the number of mixture tokens and the temperature of the softmax in the cross-attention.
    For (a), we set the attention temperature to $8$. For (b), we fix the number of mixture tokens  $K=64$. Increasing the number of mixture tokens improves downstream performance. \method{}'s performance is robust to temperature values, but a large temperature leads to a degradation in accuracy.}
    \label{fig:ablation-lalip}
\end{figure}

\textbf{Llip hyperparameters.}
\method{} introduces two hyper-parameters: the number of mixture tokens and the temperature of the softmax of the cross-attention module. In Figure~\ref{fig:ablation-lalip} we show the result of our study on both parameters conducted with a ViT-B/32.

\textbf{Number of mixtures tokens.} In Figure~\ref{fig:K-lalip}, we find that increasing the number of mixture tokens consistently improves ImageNet's top-1 accuracy without changing the model size.
Moreover, as illustrated in Figure~\ref{fig:scale}, \method{}'s performance also scales with the model size.
\method{} enables three axes to scale the model: increasing the encoder's size, decreasing image patch size or increasing the number of mixture tokens.

\textbf{Effect of softmax temperature.}
In Figure~\ref{fig:t-lalip}, we also explore the effect of the softmax temperature. The temperature controls the sharpness of the softmax's output distribution.
In each case we use the same temperature during training and inference.
Higher temperatures lead to logits with higher magnitudes leading to sharper activations. \method{} tends to be robust to a range of temperature values but its performance degrades for large temperatures.



%% file: conclusion.tex
\section{Conclusion}
In this work, we propose Llip -- a contrastive vision-language pre-training model with contextualization of visual features to model the diversity of possible captions that could match a given image.
We show that a simple approach for deriving context from the text caption and conditioning visual features leads to
richer representations and better downstream zero-shot performance on a wide variety of classifications and retrieval benchmarks.
Our detailed ablation studies show the benefits of each components of Llip and its robustness to hyperparameters.
We hope the strength of the model on downstream tasks and its simplicity will inspire the adoption of this approach in broader scenarios.

%% file: appendix.tex
\section*{Appendix}

\section{Training setup and hyperparameters}
\label{app:hypers}

We compare our training setup in Table~\ref{tab:train-protocol} where we compare the training datasets, the amount of samples seen and the batch size across the methods.~\method{} uses the same dataset as MetaCLIP and the same batch size and amount of samples seen as MetaCLIP and CLIP. Notably, it sees less samples than the other baselines and uses a smaller dataset than SigLIP.

\begin{table}[h]
\centering
\caption{Training protocol of the baselines and~\method{}: the dataset used, the number of samples seen during training and the batch size.}
\label{tab:train-protocol}
\begin{tabular}{lccc}
& data & samples seen & batch size\\
\hline
CLIP & WIT-400M & 12.8B & 32K \\
SigLIP & WebLI-10B & 40B & 32K   \\
OpenCLIP & LAION-2B & 39B & 160K\\
MetaCLIP & MetaCLIP-2.5B & 12.8B & 32K\\ 
EVA CLIP & LAION-2B & 11B+9B & 144K  \\
\method{} & MetaCLIP-2.5B & 12.8B & 32K \\
\hline
\end{tabular}
\end{table}

The hyperparameters that we used for our method are precisely the same hyper-parameters that were used for training MetaCLIP and CLIP with the only exception of the beta2 parameter of Adam set to $0.95$, the initialization of the scale  and the additional bias is -10 as in SigLIP.

For zero-shot evaluation, an image has to be encoded with the target caption. Since every targets is encoded with every images and we do not know a priori which is the right target, the ground truth target cannot leak in the prediction. To reduce the compute and memory overhead in zero-shot classification, we average the text predictions and the cross-attention queries over the template axis.

\section{Additional Results}
\label{app:additional_results}




\subsection{Robustness}
\label{app_subsec:robustness}

In Table \ref{app_tab:robustness} we show additional results on robustness benchmarks including out-of-distribution ImageNet variants across model sizes. We also show performance on geographic diversity broken down by region and model type as well as attributes from MIT States in Table \ref{app_tab:diversity}. We find while the larger \method{} model was not tuned based on the temperature parameter, when properly tuned \method{} outperforms the baselines across all DollarStreet regions with a smaller encoder.

\begin{table}[h]
\centering
\caption{\textbf{Robustness results on ViT-B/32, ViT-B/16 and ViT-L/14}.}
\label{app_tab:robustness}
\begin{tabular}{l|c|cccccc}
& Average & Val & V2 & Sketch & R & W & A \\
\hline
\textit{ViT-B/32}&&&&&&& \\
SigLIP & 57.8 & 67.3 & 59.1 & 56.2 & 76.7 & 58.4 & 28.9 \\
\rowcolor{llip}
$\text{\method{}}_{128}$    & \textbf{62.8} & 71.2 & 62.9 & 60.6 & 82.6 & 62.9 & 36.3  \\
\textit{ViT-B/16}&&&&&&& \\
SigLIP & 66.0 & 72.1 & 65.0 & 61.2 & 84.0 & 65.4 & 48.3 \\
\rowcolor{llip}
$\text{\method{}}_{64}$ & \textbf{69.7} & 75.3 & 68.3 & 63.8 & 86.6 & 69.2 & 55.0 \\
\textit{ViT-L/14}&&&&&&& \\
MetaCLIP* & 76.6 & 79.2 & 72.5 & 68.9 & 91.8 & 75.4 & 72.0 \\
\rowcolor{llip}
$\text{\method{}}_{32}$ & \textbf{79.1} & 80.9 & 74.8 & 70.5 & 93.6 & 78.0 & 76.7 \\
\hline
\end{tabular}
\end{table}

\begin{table}
\centering
\caption{\label{app_tab:diversity} \textbf{Diversity across geographies.}}
\begin{tabular}{lccccc}
                    & Africa  & Asia    & Europe  & Americas & Overall Top5            \\
\hline
\textit{ViT-B/16:} \\
MetaCLIP                & 70.38                & 80.85                & 84.12                & 82.17                     & 79.65                                   \\
SigLIP                  & 74.21                & 80.02                & 84.45                & 82.08                     & 79.94                                     \\
\rowcolor{llip}
\method{}$_{64}$                  & 74.38                & 81.26                & 85.45                & 83.17                     & 80.93                                   \\
\textit{ViT-L/14:} \\
MetaCLIP                 & 79.23                & 85.66                & 88.42                & 87.87                     & 85.26                                    \\
\rowcolor{llip}
\method{}$_{32}$                 & 76.94                & 84.44                & 86.33                & 85.61                     & 83.55                                   \\
\hline
\end{tabular}
\end{table}

\begin{table}
\centering
\caption{\label{scene_and_video_understanding}\textbf{Scene and video understanding.}  We compare MetaCLIP to \method{} on two scene understanding tasks (CLEVRCount, SUN397) and two video Understanding tasks. Both models use a ViT-L/14 encoder. While \method{} is competitive on both type of tasks, results show that the gain of \method{} are more pronounced on video understanding tasks.
MetaCLIP performance is reported from:
$^1$:~\citep{metaclip}.}
\label{tab:scenevideo_understanding}
\begin{tabular}{lccc|ccc}
& \multicolumn{3}{c|}{Scene Understanding} & \multicolumn{3}{c}{Video Understanding}   \\
 & CLEVR    & SUN397  & Avg  & KITTI  & UCF101 & Avg \\
 \hline
MetaCLIP$^1$ & \textbf{25.9} & 73.6 & 49.8 & 29.6  & 81.6  & 55.6 \\
\rowcolor{llip}
\method{}$_{32}$ & 25.5 & \textbf{74.3} & \textbf{49.9} & \textbf{34.7} & \textbf{84.5} & \textbf{59.6}  \\ 
\hline
\end{tabular}
\end{table}


\subsection{Scene and video understanding.} 
In Table~\ref{tab:scenevideo_understanding}, we focus specifically on scene and video understanding. We compare MetaCLIP to \method{} on two scene understanding tasks (CLEVRCount, SUN397) and two video understanding tasks (KITTI, UCF101).
We find the gains of \method{} are more pronounced on video understanding tasks where
the model obtains $+5.0\%$ on KITTI and $+2.8\%$ on UCF101. 

\subsection{Using image tokens in the cross-attention}
\label{app_subsec:all_tokens}
While the input to \method{}'s vision encoder is always $P$ image tokens and $K$ additional visual mixture tokens, in the standard version of \method{} we only use the outputs of the visual mixture tokens in the cross-attention (\eqref{eqn:mixing}).
In this experiment, we also included the outputs of the image patch tokens at the last layer of ViT together with the visual mixture tokens in the cross-attention (so $P + K$ tokens are used in total).

\begin{figure}[h]
    \centering
    \includegraphics[width=0.4\textwidth]{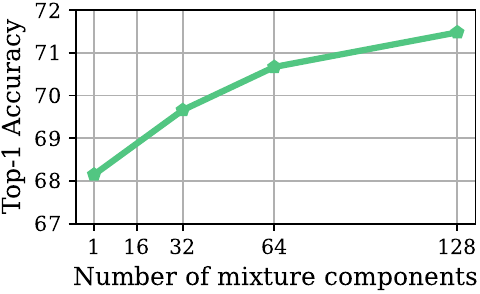}
    \caption{\textbf{Using image patch tokens together with additional visual mixture tokens in \method{}.} We report zero-shot top-1 ImageNet accuracy against the number of visual mixture tokens for a ViT-B/32 visual encoder.
    We train \method{} with temperature $\tau=1$.
    Similarly to results in Figure \ref{fig:ablation-lalip}, increasing the number of mixture tokens improves downstream performance. }
    \label{fig:all_tokens}
\end{figure}

We use \method{} with ViT-B/32 for which we have $P=49$ image patch tokens, and we report results on ImageNet zero-shot classification varying the number of visual mixture tokens $K$ in Figure \ref{fig:all_tokens}. We train the model with temperature $\tau=1$.
We can see a similar trend as in Figure \ref{fig:ablation-lalip}: the model performance increases with the higher number of the mixture tokens $K$.

Moreover, \method{} with a smaller number of additional visual mixture tokens $K=32$  (see Figure \ref{fig:ablation-lalip}) is more effective than \method{} using $P=49$ image patch tokens and $K=1$ mixture token (note that in the latter case the total number of tokens used in the cross-attention is higher, however, the number of additional mixture tokens used affects the performance more). We hypothesize that additional learnable tokens enable learning more expressive features leading to stronger performance.

\subsection{Comparison of the compute time vs accuracy of Llip with CLIP}
\begin{figure}
    \centering
    \begin{subfigure}[t]{0.48\textwidth}
    \includegraphics[width=\textwidth]{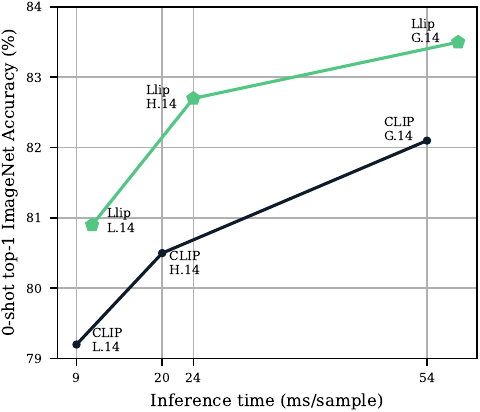}
    \caption{Zero-shot ImageNet accuracy top-1 accuracy against the inference time of inferring one ImageNet sample for vision encoders of various sizes.}
    \label{fig:scale_inference_time}
    \end{subfigure}
    \hfill
        \begin{subfigure}[t]{0.48\textwidth}
    \includegraphics[width=\textwidth]{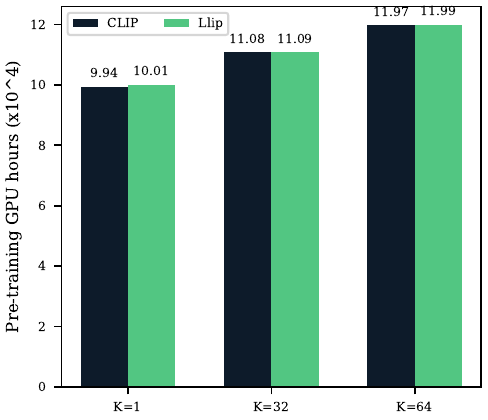}
    \caption{Effect of increasing the number of mixture tokens on the estimated amount of compute required for pre-training a ViT-G/14 backbone using the training recipe of~\citep{clip}. We find that the biggest additional cost of pre-training Llip comes from the additional mixture tokens in the vision transformer. The cost of computing the objective function is negligible. }
    \label{fig:gpu_hours_k}
    \end{subfigure}
    \caption{Analysis of the compute overhead of using Llip's contextualization for \textbf{(a)} zero-shot inference vs. ImageNet's zero-shot transfer accuracy and \textbf{(b)} estimated pre-training GPU hours of Llip compared to CLIP.}
\end{figure}

\textbf{Inference time} Figure~\ref{fig:scale} shows that the additional number of FLOPs for making an ImageNet prediction with Llip becomes marginal compared to CLIP as we scale up the encoder size. The same conclusion may be made with respect to the inference time for making an ImageNet prediction. 
In Figure~\ref{fig:scale_inference_time}, we report the inference time for IN1K's 0-shot (1000 prompts per image) Llip's inference time is slightly higher than CLIP for the same model size, while having 1.7\% improvement on 0-shot IN1K with a ViT-L/14, 2.2\% with a ViT-H/14 and 1.4\% with a ViT-G/14. Additionally Llip outperforms larger CLIP models while requiring a significantly lower inference time.

\textbf{Pre-training GPU hours} In Figure~\ref{fig:gpu_hours_k}, we present the amount of GPU hours that it takes for pre-training Llip and MetaCLIP for different number of mixture tokens. For estimating the amount of GPU hours, we compute the number of samples processed per hour on one A-100. We extrapolate the amount of samples processed per hour to obtain time it takes to process 12.8B samples.

While we see an increasing cost for pre-training Llip, this increase is not due to the objective of Llip. The cost of pre-training CLIP and Llip with the ViT-G/14 is almost identical when we fix the amount of mixture tokens processed by the vision transformer. Thus, the additional cost does not come from the contextualization per se, but the additional computation of the mixture tokens.

%% file: fair.bbl
\begin{thebibliography}{74}
\providecommand{\natexlab}[1]{#1}
\providecommand{\url}[1]{\texttt{#1}}
\expandafter\ifx\csname urlstyle\endcsname\relax
  \providecommand{\doi}[1]{doi: #1}\else
  \providecommand{\doi}{doi: \begingroup \urlstyle{rm}\Url}\fi

\bibitem[Assran et~al.(2022)Assran, Caron, Misra, Bojanowski, Bordes, Vincent, Joulin, Rabbat, and Ballas]{assran2022masked}
Assran, M., Caron, M., Misra, I., Bojanowski, P., Bordes, F., Vincent, P., Joulin, A., Rabbat, M., and Ballas, N.
\newblock Masked siamese networks for label-efficient learning.
\newblock In \emph{European Conference on Computer Vision}, pp.\  456--473. Springer, 2022.

\bibitem[Assran et~al.(2023)Assran, Duval, Misra, Bojanowski, Vincent, Rabbat, LeCun, and Ballas]{assran2023self}
Assran, M., Duval, Q., Misra, I., Bojanowski, P., Vincent, P., Rabbat, M., LeCun, Y., and Ballas, N.
\newblock Self-supervised learning from images with a joint-embedding predictive architecture.
\newblock In \emph{Proceedings of the IEEE/CVF Conference on Computer Vision and Pattern Recognition}, pp.\  15619--15629, 2023.

\bibitem[Baevski et~al.(2022)Baevski, Hsu, Xu, Babu, Gu, and Auli]{baevski2022data2vec}
Baevski, A., Hsu, W.-N., Xu, Q., Babu, A., Gu, J., and Auli, M.
\newblock Data2vec: A general framework for self-supervised learning in speech, vision and language.
\newblock \emph{arXiv preprint arXiv:2202.03555}, 2022.

\bibitem[Bardes et~al.(2024)Bardes, Garrido, Ponce, Chen, Rabbat, LeCun, Assran, and Ballas]{bardes2024revisiting}
Bardes, A., Garrido, Q., Ponce, J., Chen, X., Rabbat, M., LeCun, Y., Assran, M., and Ballas, N.
\newblock Revisiting feature prediction for learning visual representations from video.
\newblock \emph{arXiv preprint arXiv:2404.08471}, 2024.

\bibitem[Bossard et~al.(2014)Bossard, Guillaumin, and Van~Gool]{bossardFood101MiningDiscriminative2014}
Bossard, L., Guillaumin, M., and Van~Gool, L.
\newblock Food-101 {\textendash} {{Mining Discriminative Components}} with {{Random Forests}}.
\newblock In Fleet, D., Pajdla, T., Schiele, B., and Tuytelaars, T. (eds.), \emph{Computer {{Vision}} {\textendash} {{ECCV}} 2014}, Lecture {{Notes}} in {{Computer Science}}, pp.\  446--461, {Cham}, 2014. {Springer International Publishing}.
\newblock ISBN 978-3-319-10599-4.
\newblock \doi{10.1007/978-3-319-10599-4_29}.

\bibitem[Bromley et~al.(1993)Bromley, Guyon, LeCun, S{\"a}ckinger, and Shah]{bromley1993signature}
Bromley, J., Guyon, I., LeCun, Y., S{\"a}ckinger, E., and Shah, R.
\newblock Signature verification using a" siamese" time delay neural network.
\newblock \emph{Advances in neural information processing systems}, 6, 1993.

\bibitem[Chen et~al.(2020{\natexlab{a}})Chen, Kornblith, Norouzi, and Hinton]{simclr}
Chen, T., Kornblith, S., Norouzi, M., and Hinton, G.
\newblock A simple framework for contrastive learning of visual representations.
\newblock In \emph{International conference on machine learning}, pp.\  1597--1607. PMLR, 2020{\natexlab{a}}.

\bibitem[Chen et~al.(2020{\natexlab{b}})Chen, Li, Yu, El~Kholy, Ahmed, Gan, Cheng, and Liu]{chen2020uniter}
Chen, Y.-C., Li, L., Yu, L., El~Kholy, A., Ahmed, F., Gan, Z., Cheng, Y., and Liu, J.
\newblock Uniter: Universal image-text representation learning.
\newblock In \emph{European conference on computer vision}, pp.\  104--120. Springer, 2020{\natexlab{b}}.

\bibitem[Cheng et~al.(2017)Cheng, Han, and Lu]{chengRemoteSensingImage2017}
Cheng, G., Han, J., and Lu, X.
\newblock Remote {{Sensing Image Scene Classification}}: {{Benchmark}} and {{State}} of the {{Art}}.
\newblock \emph{Proceedings of the IEEE}, 105\penalty0 (10):\penalty0 1865--1883, October 2017.
\newblock ISSN 0018-9219, 1558-2256.
\newblock \doi{10.1109/JPROC.2017.2675998}.

\bibitem[Cherti et~al.(2023)Cherti, Beaumont, Wightman, Wortsman, Ilharco, Gordon, Schuhmann, Schmidt, and Jitsev]{cherti2023reproducible}
Cherti, M., Beaumont, R., Wightman, R., Wortsman, M., Ilharco, G., Gordon, C., Schuhmann, C., Schmidt, L., and Jitsev, J.
\newblock Reproducible scaling laws for contrastive language-image learning.
\newblock In \emph{Proceedings of the IEEE/CVF Conference on Computer Vision and Pattern Recognition}, pp.\  2818--2829, 2023.

\bibitem[Cimpoi et~al.(2013)Cimpoi, Maji, Kokkinos, Mohamed, and Vedaldi]{cimpoiDescribingTexturesWild2013}
Cimpoi, M., Maji, S., Kokkinos, I., Mohamed, S., and Vedaldi, A.
\newblock Describing {{Textures}} in the {{Wild}}, November 2013.

\bibitem[Coates et~al.(2010)Coates, Lee, and Ng]{coatesAnalysisSingleLayerNetworks}
Coates, A., Lee, H., and Ng, A.~Y.
\newblock An {{Analysis}} of {{Single-Layer Networks}} in {{Unsupervised Feature Learning}}, 2010.

\bibitem[Darcet et~al.(2023)Darcet, Oquab, Mairal, and Bojanowski]{darcet2023vision}
Darcet, T., Oquab, M., Mairal, J., and Bojanowski, P.
\newblock Vision transformers need registers, 2023.

\bibitem[Desai et~al.(2024)Desai, Nickel, Rajpurohit, Johnson, and Vedantam]{desai2024hyperbolic}
Desai, K., Nickel, M., Rajpurohit, T., Johnson, J., and Vedantam, R.
\newblock Hyperbolic image-text representations, 2024.

\bibitem[Dosovitskiy et~al.(2020)Dosovitskiy, Beyer, Kolesnikov, Weissenborn, Zhai, Unterthiner, Dehghani, Minderer, Heigold, Gelly, et~al.]{dosovitskiy2020image}
Dosovitskiy, A., Beyer, L., Kolesnikov, A., Weissenborn, D., Zhai, X., Unterthiner, T., Dehghani, M., Minderer, M., Heigold, G., Gelly, S., et~al.
\newblock An image is worth 16x16 words: Transformers for image recognition at scale.
\newblock \emph{arXiv preprint arXiv:2010.11929}, 2020.

\bibitem[Dou et~al.(2022)Dou, Kamath, Gan, Zhang, Wang, Li, Liu, Liu, LeCun, Peng, Gao, and Wang]{fiber}
Dou, Z.-Y., Kamath, A., Gan, Z., Zhang, P., Wang, J., Li, L., Liu, Z., Liu, C., LeCun, Y., Peng, N., Gao, J., and Wang, L.
\newblock Coarse-to-{{Fine Vision-Language Pre-training}} with {{Fusion}} in the {{Backbone}}, November 2022.

\bibitem[Fang et~al.(2024)Fang, Jose, Jain, Schmidt, Toshev, and Shankar]{dfn}
Fang, A., Jose, A.~M., Jain, A., Schmidt, L., Toshev, A.~T., and Shankar, V.
\newblock Data filtering networks.
\newblock In \emph{The Twelfth International Conference on Learning Representations}, 2024.
\newblock URL \url{https://openreview.net/forum?id=KAk6ngZ09F}.

\bibitem[Fini et~al.(2023)Fini, Astolfi, Romero-Soriano, Verbeek, and Drozdzal]{fini2023improved}
Fini, E., Astolfi, P., Romero-Soriano, A., Verbeek, J., and Drozdzal, M.
\newblock Improved baselines for vision-language pre-training.
\newblock \emph{Transactions on Machine Learning Research}, 2023.
\newblock ISSN 2835-8856.
\newblock URL \url{https://openreview.net/forum?id=a7nvXxNmdV}.
\newblock Featured Certification.

\bibitem[Foucault(1990)]{foucault1990mots}
Foucault, M.
\newblock \emph{Les mots et les choses}.
\newblock Gallimard Paris, 1990.

\bibitem[Gao et~al.(2022)Gao, Liu, Xu, Zhang, Li, Ji, and Shen]{gao2022pyramidclip}
Gao, Y., Liu, J., Xu, Z., Zhang, J., Li, K., Ji, R., and Shen, C.
\newblock Pyramidclip: Hierarchical feature alignment for vision-language model pretraining, 2022.

\bibitem[Geiger et~al.(2012)Geiger, Lenz, and Urtasun]{kitti}
Geiger, A., Lenz, P., and Urtasun, R.
\newblock Are we ready for autonomous driving? the kitti vision benchmark suite.
\newblock In \emph{2012 IEEE Conference on Computer Vision and Pattern Recognition}, pp.\  3354--3361, 2012.
\newblock \doi{10.1109/CVPR.2012.6248074}.

\bibitem[Helber et~al.(2019)Helber, Bischke, Dengel, and Borth]{helberEuroSATNovelDataset2019}
Helber, P., Bischke, B., Dengel, A., and Borth, D.
\newblock {{EuroSAT}}: {{A Novel Dataset}} and {{Deep Learning Benchmark}} for {{Land Use}} and {{Land Cover Classification}}, February 2019.

\bibitem[Hua et~al.(2021)Hua, Wang, Xue, Ren, Wang, and Zhao]{decorrelation}
Hua, T., Wang, W., Xue, Z., Ren, S., Wang, Y., and Zhao, H.
\newblock On feature decorrelation in self-supervised learning.
\newblock In \emph{2021 IEEE/CVF International Conference on Computer Vision (ICCV)}, pp.\  9578--9588, Los Alamitos, CA, USA, oct 2021. IEEE Computer Society.
\newblock \doi{10.1109/ICCV48922.2021.00946}.
\newblock URL \url{https://doi.ieeecomputersociety.org/10.1109/ICCV48922.2021.00946}.

\bibitem[Ilharco et~al.(2021)Ilharco, Wortsman, Wightman, Gordon, Carlini, Taori, Dave, Shankar, Namkoong, Miller, Hajishirzi, Farhadi, and Schmidt]{openclip}
Ilharco, G., Wortsman, M., Wightman, R., Gordon, C., Carlini, N., Taori, R., Dave, A., Shankar, V., Namkoong, H., Miller, J., Hajishirzi, H., Farhadi, A., and Schmidt, L.
\newblock Openclip, July 2021.
\newblock URL \url{https://doi.org/10.5281/zenodo.5143773}.
\newblock If you use this software, please cite it as below.

\bibitem[Isola et~al.(2015)Isola, Lim, and Adelson]{mit_states}
Isola, P., Lim, J.~J., and Adelson, E.~H.
\newblock Discovering states and transformations in image collections.
\newblock In \emph{Proceedings of the IEEE Conference on Computer Vision and Pattern Recognition (CVPR)}, June 2015.

\bibitem[Jia et~al.(2021)Jia, Yang, Xia, Chen, Parekh, Pham, Le, Sung, Li, and Duerig]{jia2021scaling}
Jia, C., Yang, Y., Xia, Y., Chen, Y.-T., Parekh, Z., Pham, H., Le, Q.~V., Sung, Y., Li, Z., and Duerig, T.
\newblock Scaling up visual and vision-language representation learning with noisy text supervision, 2021.

\bibitem[Jing et~al.(2022)Jing, Vincent, LeCun, and Tian]{jing2022understanding}
Jing, L., Vincent, P., LeCun, Y., and Tian, Y.
\newblock Understanding dimensional collapse in contrastive self-supervised learning.
\newblock In \emph{International Conference on Learning Representations}, 2022.
\newblock URL \url{https://openreview.net/forum?id=YevsQ05DEN7}.

\bibitem[Kingma \& Ba(2017)Kingma and Ba]{kingma2017adam}
Kingma, D.~P. and Ba, J.
\newblock Adam: A method for stochastic optimization, 2017.

\bibitem[Krause et~al.(2013)Krause, Deng, Stark, and {Fei-Fei}]{krauseCollectingLargeScaleDataset}
Krause, J., Deng, J., Stark, M., and {Fei-Fei}, L.
\newblock Collecting a {{Large-Scale Dataset}} of {{Fine-Grained Cars}}, 2013.

\bibitem[Krizhevsky(2010)]{krizhevskyLearningMultipleLayers}
Krizhevsky, A.
\newblock Learning {{Multiple Layers}} of {{Features}} from {{Tiny Images}}, 2010.

\bibitem[LeCun(2022)]{lecun2022path}
LeCun, Y.
\newblock A path towards autonomous machine intelligence version 0.9. 2, 2022-06-27, 2022.

\bibitem[LeCun \& Cortes(2010)LeCun and Cortes]{lecun-mnisthandwrittendigit-2010}
LeCun, Y. and Cortes, C.
\newblock {MNIST} handwritten digit database.
\newblock http://yann.lecun.com/exdb/mnist/, 2010.
\newblock URL \url{http://yann.lecun.com/exdb/mnist/}.

\bibitem[Li et~al.(2003)Li, Andreetto, and Ranzato]{CaltechUCSDBirds2002011Dataset}
Li, F.-F., Andreetto, M., and Ranzato, M.~A.
\newblock The {{Caltech-UCSD Birds-200-2011 Dataset}}.
\newblock https://authors.library.caltech.edu/records/cvm3y-5hh21, 2003.

\bibitem[Li et~al.(2021)Li, Selvaraju, Gotmare, Joty, Xiong, and Hoi]{li2021align}
Li, J., Selvaraju, R.~R., Gotmare, A.~D., Joty, S., Xiong, C., and Hoi, S.
\newblock Align before fuse: Vision and language representation learning with momentum distillation, 2021.

\bibitem[Li et~al.(2022{\natexlab{a}})Li, Li, Xiong, and Hoi]{blip}
Li, J., Li, D., Xiong, C., and Hoi, S.
\newblock {{BLIP}}: {{Bootstrapping Language-Image Pre-training}} for {{Unified Vision-Language Understanding}} and {{Generation}}, February 2022{\natexlab{a}}.

\bibitem[Li et~al.(2023{\natexlab{a}})Li, Li, Savarese, and Hoi]{blip2}
Li, J., Li, D., Savarese, S., and Hoi, S.
\newblock {{BLIP-2}}: {{Bootstrapping Language-Image Pre-training}} with {{Frozen Image Encoders}} and {{Large Language Models}}, June 2023{\natexlab{a}}.

\bibitem[Li et~al.(2023{\natexlab{b}})Li, Li, Savarese, and Hoi]{li2023blip}
Li, J., Li, D., Savarese, S., and Hoi, S.
\newblock Blip-2: Bootstrapping language-image pre-training with frozen image encoders and large language models.
\newblock \emph{arXiv preprint arXiv:2301.12597}, 2023{\natexlab{b}}.

\bibitem[Li et~al.(2023{\natexlab{c}})Li, Li, Savarese, and Hoi]{li2023blip2}
Li, J., Li, D., Savarese, S., and Hoi, S.
\newblock Blip-2: Bootstrapping language-image pre-training with frozen image encoders and large language models, 2023{\natexlab{c}}.

\bibitem[Li et~al.(2022{\natexlab{b}})Li, Zhang, Zhang, Yang, Li, Zhong, Wang, Yuan, Zhang, Hwang, Chang, and Gao]{li2022grounded}
Li, L.~H., Zhang, P., Zhang, H., Yang, J., Li, C., Zhong, Y., Wang, L., Yuan, L., Zhang, L., Hwang, J.-N., Chang, K.-W., and Gao, J.
\newblock Grounded language-image pre-training, 2022{\natexlab{b}}.

\bibitem[Li et~al.(2020)Li, Gao, Niu, Xiao, Liu, Liu, Wu, and Wang]{li2020unimo}
Li, W., Gao, C., Niu, G., Xiao, X., Liu, H., Liu, J., Wu, H., and Wang, H.
\newblock Unimo: Towards unified-modal understanding and generation via cross-modal contrastive learning.
\newblock \emph{arXiv preprint arXiv:2012.15409}, 2020.

\bibitem[Li et~al.(2023{\natexlab{d}})Li, Wang, and Xie]{li2023clipav2}
Li, X., Wang, Z., and Xie, C.
\newblock Clipa-v2: Scaling clip training with 81.1

\bibitem[Li et~al.(2023{\natexlab{e}})Li, Fan, Hu, Feichtenhofer, and He]{flip}
Li, Y., Fan, H., Hu, R., Feichtenhofer, C., and He, K.
\newblock Scaling language-image pre-training via masking.
\newblock In \emph{Proceedings of the IEEE/CVF Conference on Computer Vision and Pattern Recognition}, pp.\  23390--23400, 2023{\natexlab{e}}.

\bibitem[Lin et~al.(2014)Lin, Maire, Belongie, Hays, Perona, Ramanan, Doll{\'a}r, and Zitnick]{mscoco}
Lin, T.-Y., Maire, M., Belongie, S., Hays, J., Perona, P., Ramanan, D., Doll{\'a}r, P., and Zitnick, C.~L.
\newblock Microsoft coco: Common objects in context.
\newblock In Fleet, D., Pajdla, T., Schiele, B., and Tuytelaars, T. (eds.), \emph{Computer Vision -- ECCV 2014}, pp.\  740--755, Cham, 2014. Springer International Publishing.
\newblock ISBN 978-3-319-10602-1.

\bibitem[Liu et~al.(2023)Liu, Li, Li, and Lee]{liu2023improved}
Liu, H., Li, C., Li, Y., and Lee, Y.~J.
\newblock Improved baselines with visual instruction tuning, 2023.

\bibitem[Loshchilov \& Hutter(2017)Loshchilov and Hutter]{loshchilov2017decoupled}
Loshchilov, I. and Hutter, F.
\newblock Decoupled weight decay regularization.
\newblock \emph{arXiv preprint arXiv:1711.05101}, 2017.

\bibitem[Maji et~al.(2013)Maji, Rahtu, Kannala, Blaschko, and Vedaldi]{majiFineGrainedVisualClassification2013}
Maji, S., Rahtu, E., Kannala, J., Blaschko, M., and Vedaldi, A.
\newblock Fine-{{Grained Visual Classification}} of {{Aircraft}}, June 2013.

\bibitem[Misra \& van~der Maaten(2020)Misra and van~der Maaten]{pirl2020}
Misra, I. and van~der Maaten, L.
\newblock Self-supervised learning of pretext-invariant representations.
\newblock In \emph{2020 IEEE/CVF Conference on Computer Vision and Pattern Recognition (CVPR)}, pp.\  6706--6716, Los Alamitos, CA, USA, jun 2020. IEEE Computer Society.
\newblock \doi{10.1109/CVPR42600.2020.00674}.
\newblock URL \url{https://doi.ieeecomputersociety.org/10.1109/CVPR42600.2020.00674}.

\bibitem[Moayeri et~al.(2023)Moayeri, Rezaei, Sanjabi, and Feizi]{moayeri2023texttoconcept}
Moayeri, M., Rezaei, K., Sanjabi, M., and Feizi, S.
\newblock Text-to-concept (and back) via cross-model alignment, 2023.

\bibitem[Moon et~al.(2023)Moon, Madotto, Lin, Nagarajan, Smith, Jain, Yeh, Murugesan, Heidari, Liu, Srinet, Damavandi, and Kumar]{moon2023anymal}
Moon, S., Madotto, A., Lin, Z., Nagarajan, T., Smith, M., Jain, S., Yeh, C.-F., Murugesan, P., Heidari, P., Liu, Y., Srinet, K., Damavandi, B., and Kumar, A.
\newblock Anymal: An efficient and scalable any-modality augmented language model, 2023.

\bibitem[Mu et~al.(2021)Mu, Kirillov, Wagner, and Xie]{mu2021slip}
Mu, N., Kirillov, A., Wagner, D., and Xie, S.
\newblock Slip: Self-supervision meets language-image pre-training, 2021.

\bibitem[Nilsback \& Zisserman(2008)Nilsback and Zisserman]{nilsbackAutomatedFlowerClassification2008}
Nilsback, M.-E. and Zisserman, A.
\newblock Automated {{Flower Classification}} over a {{Large Number}} of {{Classes}}.
\newblock In \emph{2008 {{Sixth Indian Conference}} on {{Computer Vision}}, {{Graphics}} \& {{Image Processing}}}, pp.\  722--729, {Bhubaneswar, India}, December 2008. {IEEE}.
\newblock \doi{10.1109/ICVGIP.2008.47}.

\bibitem[Oquab et~al.(2023)Oquab, Darcet, Moutakanni, Vo, Szafraniec, Khalidov, Fernandez, Haziza, Massa, El-Nouby, et~al.]{oquab2023dinov2}
Oquab, M., Darcet, T., Moutakanni, T., Vo, H., Szafraniec, M., Khalidov, V., Fernandez, P., Haziza, D., Massa, F., El-Nouby, A., et~al.
\newblock Dinov2: Learning robust visual features without supervision.
\newblock \emph{arXiv preprint arXiv:2304.07193}, 2023.

\bibitem[Parkhi et~al.(2012)Parkhi, Vedaldi, Zisserman, and Jawahar]{parkhiCatsDogs2012}
Parkhi, O.~M., Vedaldi, A., Zisserman, A., and Jawahar, C.~V.
\newblock Cats and dogs.
\newblock In \emph{2012 {{IEEE Conference}} on {{Computer Vision}} and {{Pattern Recognition}}}, pp.\  3498--3505, June 2012.
\newblock \doi{10.1109/CVPR.2012.6248092}.

\bibitem[Paszke et~al.(2019)Paszke, Gross, Massa, Lerer, Bradbury, Chanan, Killeen, Lin, Gimelshein, Antiga, Desmaison, Kopf, Yang, DeVito, Raison, Tejani, Chilamkurthy, Steiner, Fang, Bai, and Chintala]{pytorch}
Paszke, A., Gross, S., Massa, F., Lerer, A., Bradbury, J., Chanan, G., Killeen, T., Lin, Z., Gimelshein, N., Antiga, L., Desmaison, A., Kopf, A., Yang, E., DeVito, Z., Raison, M., Tejani, A., Chilamkurthy, S., Steiner, B., Fang, L., Bai, J., and Chintala, S.
\newblock Pytorch: An imperative style, high-performance deep learning library.
\newblock In \emph{Advances in Neural Information Processing Systems 32}, pp.\  8024--8035. Curran Associates, Inc., 2019.

\bibitem[Purushwalkam \& Gupta(2020)Purushwalkam and Gupta]{demystifying2020}
Purushwalkam, S. and Gupta, A.
\newblock Demystifying contrastive self-supervised learning: Invariances, augmentations and dataset biases.
\newblock \emph{CoRR}, abs/2007.13916, 2020.
\newblock URL \url{https://arxiv.org/abs/2007.13916}.

\bibitem[Radford et~al.(2021)Radford, Kim, Hallacy, Ramesh, Goh, Agarwal, Sastry, Askell, Mishkin, Clark, Krueger, and Sutskever]{clip}
Radford, A., Kim, J.~W., Hallacy, C., Ramesh, A., Goh, G., Agarwal, S., Sastry, G., Askell, A., Mishkin, P., Clark, J., Krueger, G., and Sutskever, I.
\newblock Learning {{Transferable Visual Models From Natural Language Supervision}}, February 2021.

\bibitem[Ramesh et~al.(2021)Ramesh, Pavlov, Goh, Gray, Voss, Radford, Chen, and Sutskever]{ramesh2021zero}
Ramesh, A., Pavlov, M., Goh, G., Gray, S., Voss, C., Radford, A., Chen, M., and Sutskever, I.
\newblock Zero-shot text-to-image generation.
\newblock In \emph{International Conference on Machine Learning}, pp.\  8821--8831. PMLR, 2021.

\bibitem[Recht et~al.(2019)Recht, Roelofs, Schmidt, and Shankar]{recht2019imagenet}
Recht, B., Roelofs, R., Schmidt, L., and Shankar, V.
\newblock Do imagenet classifiers generalize to imagenet?
\newblock In \emph{International conference on machine learning}, pp.\  5389--5400. PMLR, 2019.

\bibitem[Russakovsky et~al.(2015)Russakovsky, Deng, Su, Krause, Satheesh, Ma, Huang, Karpathy, Khosla, Bernstein, Berg, and {Fei-Fei}]{russakovskyImageNetLargeScale2015}
Russakovsky, O., Deng, J., Su, H., Krause, J., Satheesh, S., Ma, S., Huang, Z., Karpathy, A., Khosla, A., Bernstein, M., Berg, A.~C., and {Fei-Fei}, L.
\newblock {{ImageNet Large Scale Visual Recognition Challenge}}, January 2015.

\bibitem[Soomro et~al.(2012)Soomro, Zamir, and Shah]{soomroucf101}
Soomro, K., Zamir, A.~R., and Shah, M.
\newblock {UCF101:} {A} dataset of 101 human actions classes from videos in the wild.
\newblock \emph{CoRR}, abs/1212.0402, 2012.
\newblock URL \url{http://arxiv.org/abs/1212.0402}.

\bibitem[Stallkamp et~al.(2011)Stallkamp, Schlipsing, Salmen, and Igel]{Stallkamp-IJCNN-2011}
Stallkamp, J., Schlipsing, M., Salmen, J., and Igel, C.
\newblock The {G}erman {T}raffic {S}ign {R}ecognition {B}enchmark: A multi-class classification competition.
\newblock In \emph{IEEE International Joint Conference on Neural Networks}, pp.\  1453--1460, 2011.

\bibitem[Sun et~al.(2023)Sun, Fang, Wu, Wang, and Cao]{sun2023evaclip}
Sun, Q., Fang, Y., Wu, L., Wang, X., and Cao, Y.
\newblock Eva-clip: Improved training techniques for clip at scale, 2023.

\bibitem[Vaswani et~al.(2023)Vaswani, Shazeer, Parmar, Uszkoreit, Jones, Gomez, Kaiser, and Polosukhin]{vaswani2023attention}
Vaswani, A., Shazeer, N., Parmar, N., Uszkoreit, J., Jones, L., Gomez, A.~N., Kaiser, L., and Polosukhin, I.
\newblock Attention is all you need, 2023.

\bibitem[Wang et~al.(2022{\natexlab{a}})Wang, Yang, Men, Lin, Bai, Li, Ma, Zhou, Zhou, and Yang]{wang2022ofa}
Wang, P., Yang, A., Men, R., Lin, J., Bai, S., Li, Z., Ma, J., Zhou, C., Zhou, J., and Yang, H.
\newblock Ofa: Unifying architectures, tasks, and modalities through a simple sequence-to-sequence learning framework, 2022{\natexlab{a}}.

\bibitem[Wang et~al.(2022{\natexlab{b}})Wang, Yu, Yu, Dai, Tsvetkov, and Cao]{wang2022simvlm}
Wang, Z., Yu, J., Yu, A.~W., Dai, Z., Tsvetkov, Y., and Cao, Y.
\newblock Sim{VLM}: Simple visual language model pretraining with weak supervision.
\newblock In \emph{International Conference on Learning Representations}, 2022{\natexlab{b}}.
\newblock URL \url{https://openreview.net/forum?id=GUrhfTuf_3}.

\bibitem[Xiao et~al.(2010)Xiao, Hays, Ehinger, Oliva, and Torralba]{xiaoSUNDatabaseLargescale2010}
Xiao, J., Hays, J., Ehinger, K.~A., Oliva, A., and Torralba, A.
\newblock {{SUN}} database: {{Large-scale}} scene recognition from abbey to zoo.
\newblock In \emph{2010 {{IEEE Computer Society Conference}} on {{Computer Vision}} and {{Pattern Recognition}}}, pp.\  3485--3492, June 2010.
\newblock \doi{10.1109/CVPR.2010.5539970}.

\bibitem[Xu et~al.(2023)Xu, Xie, Tan, Huang, Howes, Sharma, Li, Ghosh, Zettlemoyer, and Feichtenhofer]{metaclip}
Xu, H., Xie, S., Tan, X.~E., Huang, P.-Y., Howes, R., Sharma, V., Li, S.-W., Ghosh, G., Zettlemoyer, L., and Feichtenhofer, C.
\newblock Demystifying {{CLIP Data}}, October 2023.

\bibitem[Ye et~al.(2020)Ye, Yao, Xue, and Li]{Ye2020WeaklySL}
Ye, W., Yao, J., Xue, H., and Li, Y.
\newblock Weakly supervised lesion localization with probabilistic-cam pooling.
\newblock \emph{ArXiv}, abs/2005.14480, 2020.
\newblock URL \url{https://api.semanticscholar.org/CorpusID:215776849}.

\bibitem[Young et~al.(2014)Young, Lai, Hodosh, and Hockenmaier]{flickr}
Young, P., Lai, A., Hodosh, M., and Hockenmaier, J.
\newblock From image descriptions to visual denotations: New similarity metrics for semantic inference over event descriptions.
\newblock \emph{Transactions of the Association for Computational Linguistics}, 2:\penalty0 67--78, 2014.
\newblock \doi{10.1162/tacl_a_00166}.
\newblock URL \url{https://aclanthology.org/Q14-1006}.

\bibitem[Yu et~al.(2022)Yu, Wang, Vasudevan, Yeung, Seyedhosseini, and Wu]{yu2022coca}
Yu, J., Wang, Z., Vasudevan, V., Yeung, L., Seyedhosseini, M., and Wu, Y.
\newblock Coca: Contrastive captioners are image-text foundation models, 2022.

\bibitem[Zhai et~al.(2022)Zhai, Wang, Mustafa, Steiner, Keysers, Kolesnikov, and Beyer]{zhai2022lit}
Zhai, X., Wang, X., Mustafa, B., Steiner, A., Keysers, D., Kolesnikov, A., and Beyer, L.
\newblock Lit: Zero-shot transfer with locked-image text tuning, 2022.

\bibitem[Zhai et~al.(2023)Zhai, Mustafa, Kolesnikov, and Beyer]{siglip}
Zhai, X., Mustafa, B., Kolesnikov, A., and Beyer, L.
\newblock Sigmoid {{Loss}} for {{Language Image Pre-Training}}, September 2023.

\bibitem[Zhang et~al.(2021)Zhang, Li, Hu, Yang, Zhang, Wang, Choi, and Gao]{Zhang_2021_CVPR}
Zhang, P., Li, X., Hu, X., Yang, J., Zhang, L., Wang, L., Choi, Y., and Gao, J.
\newblock Vinvl: Revisiting visual representations in vision-language models.
\newblock In \emph{Proceedings of the IEEE/CVF Conference on Computer Vision and Pattern Recognition (CVPR)}, pp.\  5579--5588, June 2021.

\bibitem[Zhou et~al.(2022)Zhou, Yang, Loy, and Liu]{Zhou_2022_CVPR}
Zhou, K., Yang, J., Loy, C.~C., and Liu, Z.
\newblock Conditional prompt learning for vision-language models.
\newblock In \emph{Proceedings of the IEEE/CVF Conference on Computer Vision and Pattern Recognition (CVPR)}, pp.\  16816--16825, June 2022.

\end{thebibliography}


\begin{thebibliography}{0}
\providecommand{\natexlab}[1]{#1}
\providecommand{\url}[1]{\texttt{#1}}
\expandafter\ifx\csname urlstyle\endcsname\relax
  \providecommand{\doi}[1]{doi: #1}\else
  \providecommand{\doi}{doi: \begingroup \urlstyle{rm}\Url}\fi

\end{thebibliography}
